\documentclass[twoside,11pt]{article}

\usepackage{blindtext}
\usepackage{dsfont}

\usepackage[preprint]{jmlr2e}

\usepackage{hyperref}
\usepackage[dvipsnames]{xcolor}
\usepackage{amsmath}
\usepackage{mathtools}

\usepackage{mathrsfs}
\usepackage[capitalize,noabbrev]{cleveref}

\usepackage{enumitem}
\usepackage{siunitx}

\usepackage{comment}

\newtheorem{cor}[theorem]{Corollary}
\newtheorem{prop}[theorem]{Proposition}

\usepackage{dirtytalk}

\usepackage[textsize=tiny]{todonotes}
\usepackage{regexpatch}
\makeatletter
\xpatchcmd{\@todo}{\setkeys{todonotes}{#1}}{\setkeys{todonotes}{inline,#1}}{}{}
\makeatother

\usepackage{lastpage}
\jmlrheading{27}{2026+}{1--\pageref{LastPage} \textit{under review}}{5/25; Revised 12/26}{?/?}{21-0000}{Julian Rodemann and James Bailie}

\DeclareMathOperator*{\argmin}{arg\,min}

\newcommand{\iid}{i.i.d.}
\renewcommand{\varepsilon}{\epsilon} 
\newcommand{\iidsim}{\stackrel{\iid}{\sim}}

\newcommand{\diamZ}{\mathscr{D}_{\mathcal Z}}

\newcommand{\diamY}{\mathscr{D}_{\mathcal Y}}
\newcommand{\dz}{d_{z}}
\newcommand{\dx}{d_{x}}
\newcommand{\dy}{d_{y}}

\newcounter{regularityConditionLabel}[condition]

\newcommand{\regularityConditionLabel}[1]{%
    \begingroup%
        \refstepcounter{regularityConditionLabel}%
        \label{#1}%
    \endgroup%
}

\title{Generalization Bounds and Stopping Rules for Learning with Adaptively Selected Data}

\ShortHeadings{Generalizing From Adaptively Selected Data}{Rodemann and Bailie}
\firstpageno{1}

\author{\name Julian Rodemann \email j.rodemann@lmu.de \\
       \addr Department of Statistics\\
       Ludwig-Maximilians-Universit\"at M\"unchen \\
       Munich, 80539, Germany
       \AND
       \name James Bailie %
        \email bailie@chalmers.edu \\
        \addr Division of Data Science and AI, Department of Computer Science and Engineering \\
        Chalmers University of Technology and the University of Gothenburg \\
        Gothenburg, 41296, Sweden}

\editor{Samuel Kaski}

\begin{document}

\maketitle

\begin{abstract}
Many learning methods, including active learning, semi-supervised learning, bandits, and boosting, adaptively select training data in light of previously learned parameters. 
This article addresses the question of how well these methods can generalize from their adaptively selected samples. In particular, we prove universal generalization bounds for this class of methods using covering numbers and Wasserstein ambiguity sets. %
Our results do not rely on any assumptions about the (often unknown) distribution of adaptively selected data; they only require the initial sample to be \iid, along with some verifiable conditions on how these methods are implemented. 
We prove results for both convergent and finite-iteration solutions. %
The latter are anytime valid, thereby giving rise to stopping rules for a practitioner seeking to guarantee the out-of-sample performance of their learning algorithm. %
We illustrate our bounds and stopping rules for the special case of semi-supervised learning. %
\end{abstract}

\begin{keywords}
  Reciprocal Learning, Generalization Bounds, Statistical Learning Theory, Semi-Supervised Learning, Active Learning, Boosting, Bandits, Adaptively Selected Data
\end{keywords}

\section{Introduction}\label{sec:intro}

Almost eight decades after \citet{wiener1948cybernetics} pioneered the study of circular causal and feedback mechanisms (known as ``cybernetics'') \citep{von1952cybernetics}, machine learning has seen a revival of his ideas regarding stability, adaptability and feedback-driven control in recent years.\footnote{For some historical background on why it took so long for these research trends to resurface, we refer the interested reader to Appendix~\ref{app-history}.}
This revival has been driven by the insight that many real-world learning problems are inherently dynamic, involving ongoing interaction between predictive models and the systems they influence. 
One prominent manifestation of this reemerging perspective is performative prediction \citep{perdomo2020performative}, which models how predictions can affect the population on which they are made.
Its example applications range from self-negating traffic route predictions to self-fulfilling credit scores. Extensive research has emerged on understanding the population dynamics of such feedback loops \citep{miller2021outside,perf-power,brown2022performative,xue2024distributionally,perdomo2025revisiting}. %

The revival of feedback-driven perspectives, however, is not restricted to the population level. \citet{rodemann24reciprocal} observe that several learning methods like active learning, semi-supervised learning, boosting or bandit algorithms adapt the \textit{sample}---not the population---in response to the model fit. In other words, learning goes both ways in the sense that these methods share a feedback-loop structure. Not only do they iteratively learn parameters from data, but also (partially) the other
way around: The new data is a function of the previous parameters (and the previous data). In this way, there is a kind of reciprocal relationship between parameters and data in these models. \citet{rodemann24reciprocal}
generalize methods with this type of relationship into a unified framework they term \textit{reciprocal learning}.\footnote{Throughout the paper, we refer to the framework's special cases like active learning, boosting or bandits as \textit{learning paradigms} or \textit{methods}, each one in turn comprising multiple \textit{learning algorithms}, \textit{learners} or \textit{models}. On a different note, it should be added that \citet{breiman1999prediction} already recognized boosting's character as a sequential \say{prediction game,} akin to the embedding of reciprocal learning into sequential decision theory \citep[see Section~2.2 of][]{rodemann24reciprocal}.} 
(Although this name does not fully encapsulate all the dynamics of such methods,\footnote{In particular, it is questionable to what extent data can be cast as a learning entity in the same way parameters can be. Nevertheless, the overall learning process in these methods entails not only adaptivity of a learner’s parameters from data, but also adaptivity of the learner's data from parameters.} 
we will follow their terminology when referring to this framework in order %
to maintain consistency with existing literature.)

Regardless of the terminology we adopt, the immediate concern at this point in the paper is to develop an understanding of the sample--parameter feedback structure at the center of this framework.
Let us illustrate the core idea %
through a straightforward example involving one of the framework's special cases: self-training within semi-supervised learning \citep[see e.g.,][]{arazo2020pseudo, rizve2020defense, li2020pseudo, bordini2024self, dietrich2024semi}. 
The objective of semi-supervised learning is to leverage both labeled and unlabeled data in finding a predictive classification function $\hat y(\theta,x)$ parameterized by $\theta$. Self-training algorithms begin by applying empirical risk minimization (ERM) to fit a model using labeled data, subsequently employing this initial model to predict labels for unlabeled data. In the next phase, instances from the unlabeled data are chosen---guided by a \say{confidence score,} which typically measures predictive uncertainty---and added to the training set along with their predicted labels, which are called \say{pseudo-labels.} %
This process is then repeated, so that the pseudo-labeled samples at iteration $t+1$ depend on the model through its predictions from iteration $t$.
Because the new labels are predicted by the current model $\hat y(\theta_t,x)$, they depend on the parameter $\theta_t$ learned in the previous step. This dependency exemplifies how the sample iteratively adapts to the learned parameters, a process which we call the model's \textit{sample adaptation function} (Definition~\ref{def:basic-reciprocal-learning}). 

Another example of this type of sample--parameter feedback is given by boosting algorithms. Consider gradient boosting regression with squared loss for ease of exposition. Here, gradient boosting’s pseudo-residuals correspond to the residuals $y- \hat y$. 
At iteration $t$, gradient boosting thus fits a model to the residuals $y - \hat y(\theta_{t-1},x)$ from the previous iteration $t-1$. It is thus easy to see that the sample in $t$ can be expressed as a function of the sample in $t-1$ and the location shift caused by the model from $t-1$ via its prediction $\hat{y}(\theta_{t-1},x)$.

\section{Outline}\label{sec:outline}

\begin{figure}
\begin{minipage}{0.5\linewidth}
    \centering
    \includegraphics[width=0.5\linewidth]{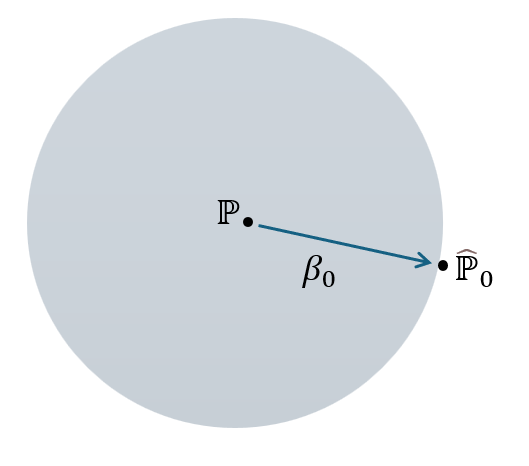}
\end{minipage}
\begin{minipage}{0.5\linewidth}
    \centering
        \includegraphics[width=0.7\linewidth]{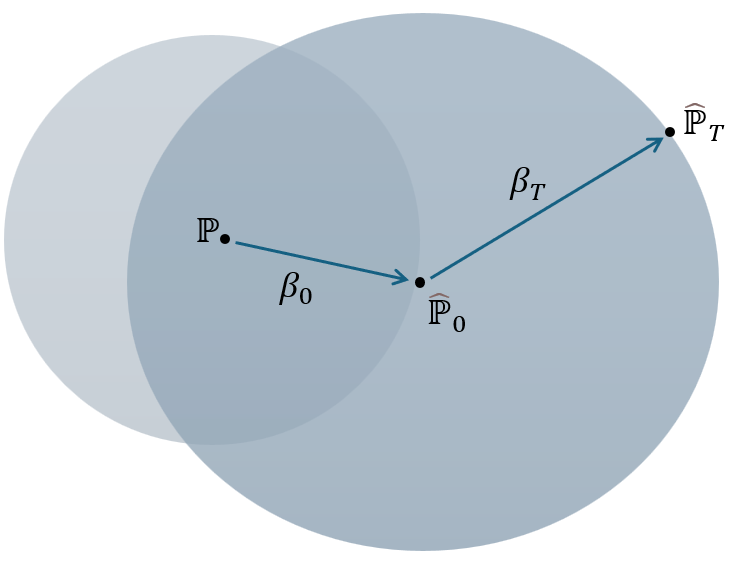}
\end{minipage}
        \caption{\textbf{Left:} Bound on $p$-Wasserstein distance $W_p$ between law $\mathbb P$ and initial \iid~sample $\hat{\mathbb P}_0$ : $W_{p}(\mathbb P,\hat{\mathbb P}_0) \leq \beta_0$ \citep{fournier2015rate}.
        \textbf{Right:} Reciprocal distortion bound between initial \iid~sample $\hat{\mathbb P}_0$ and sample $\hat{\mathbb P}_T$ in reciprocal learning iteration $T$:  $W_p(\hat{\mathbb P}_0, \hat{\mathbb P}_{T}) \leq \beta_T$ (Lemma~\ref{lemma:recipr-distortion}, see below). %
        }
        \label{fig:illu-wasserstein-space}
\end{figure}

While \citet{rodemann24reciprocal} prove reciprocal learning algorithms converge to an in-sample optimum under sufficient conditions (see their Theorem~4), they do not address the question of how these algorithms generalize on unseen test data. Their notion of optimality refers to \textit{training} optimality only (see their Definition 9). %
However, if an algorithm can iteratively change the data---and hence change the problem the algorithm aims to optimize---then convergence to an optimal solution comes as little surprise. The question of whether reciprocal learning can \textit{generalize} appears considerably harder, given the fact that reciprocal learning entails changing the sample, potentially shifting it away from the population, as we illustrate in Figure~\ref{fig:illu-wasserstein-space}.

The article at hand answers this question. We prove PAC-style generalization bounds for learning with adaptively selected data. (Here, we use the term ``PAC-style'' to indicate that our bounds provide finite-sample,
high-probability guarantees on the generalization gap and excess risk.) Specifically, we give two sets of bounds: one set regarding a reciprocal learning algorithm's generalization gap and another regarding its excess risk. The former bounds %
the difference between the training error a %
learner 
achieves and its (unknown) generalization error (Theorems~\ref{thm:gen-error-bound},~\ref{thm:gen-error-bound-anytime}). The latter bounds %
the difference between the %
learner's risk and the (unknown) minimum risk from the best learner in the hypothesis class (Theorems~\ref{thm:gen-error-bound-data-dependent},~\ref{thm:excess-risk-classic},~\ref{thm:excess-risk-classic-anytime},~\ref{thm:data-dependent-anytime-excess}). Our bounds require only generic assumptions on the model's sample adaptation and loss functions, and thus they apply to the full set of exemplar reciprocal learning methods listed in \citet{rodemann24reciprocal}.
Conceptually, the key behind our results is a reinterpretation of Wasserstein ambiguity sets. 
While typically utilized to robustify ERM with respect to ambiguity about the distributional \textit{assumption} \citep[see e.g.,][]{shafieezadeh2015distributionally}, we use these sets to analyze shifts in the \textit{empirical} distribution caused by the reciprocal learning algorithms themselves. This is an exciting change of perspective, since we have precise knowledge about this shift from the algorithm and can specify the ambiguity sets accordingly rather than having to \say{judiciously} \citep[page 1576]{shafieezadeh2015distributionally} choose them.

As illustrated by Figure~\ref{fig:illu-wasserstein-space}, we shall start by exploiting a classical bound $\beta_0$ on the \mbox{$p$-Wasserstein} distance between the law $\mathbb P$ and an initial \iid~sample $\hat{\mathbb P}_0$ thereof. Importantly, $\mathbb P$ should be understood as the \emph{target population law} underlying the
initial sample $\hat{\mathbb P}_0$, not as the law governing the entire adaptive data sequence. Our
framework starts from an initial sample $Z_1,\dots,Z_n \iidsim \mathbb P$ and then studies
what happens when this sample is subsequently modified by the learning algorithm itself.
Thus, the later samples $\hat{\mathbb P}_t$ (for $t = 1, \ldots, T$, where $T$ is the iteration at which the learning algorithm stops) need not be i.i.d.\ from $\mathbb P$ and
indeed will typically differ from it. Nevertheless, our inference target is $\mathbb P$, not the distribution from which $\hat{\mathbb P}_T$ is \iid. (Indeed, such a distribution need not exist---and in general does not exist---for the settings we consider.) This work therefore covers cases where inference about $\mathbb P$ remains meaningful across iterations; we do not intend to address inference about a moving target law.

\begin{figure}
\begin{minipage}{0.45\linewidth}
        \includegraphics[width=1.05\linewidth]{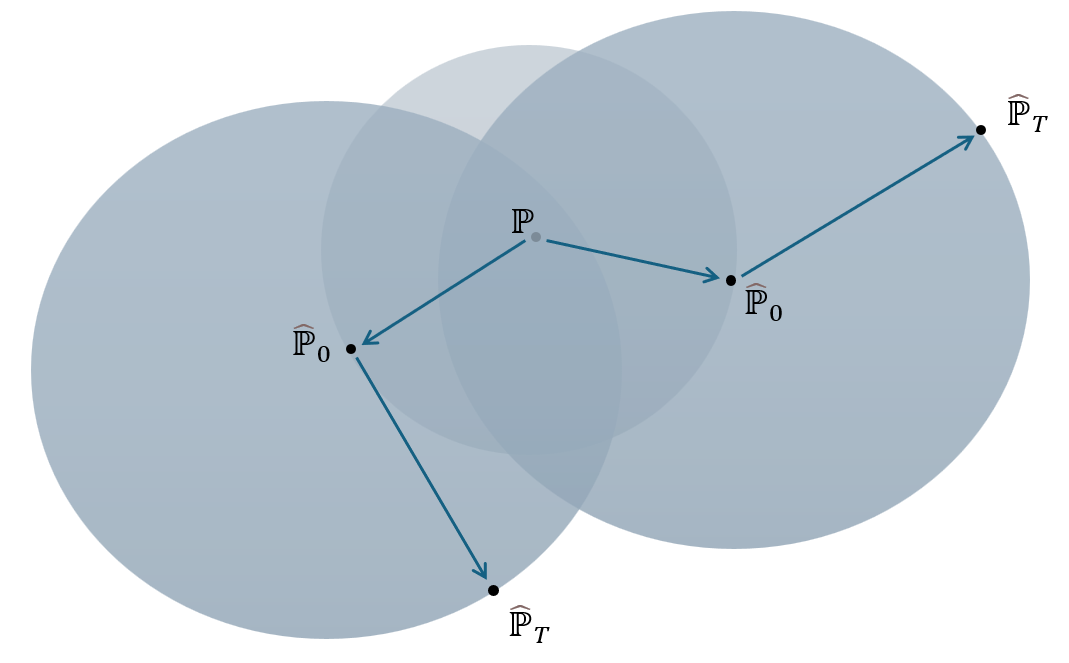}
\end{minipage}
\begin{minipage}{0.05\linewidth}    
\end{minipage}
\begin{minipage}{0.5\linewidth}
    \centering
        \includegraphics[width=1.1\linewidth]{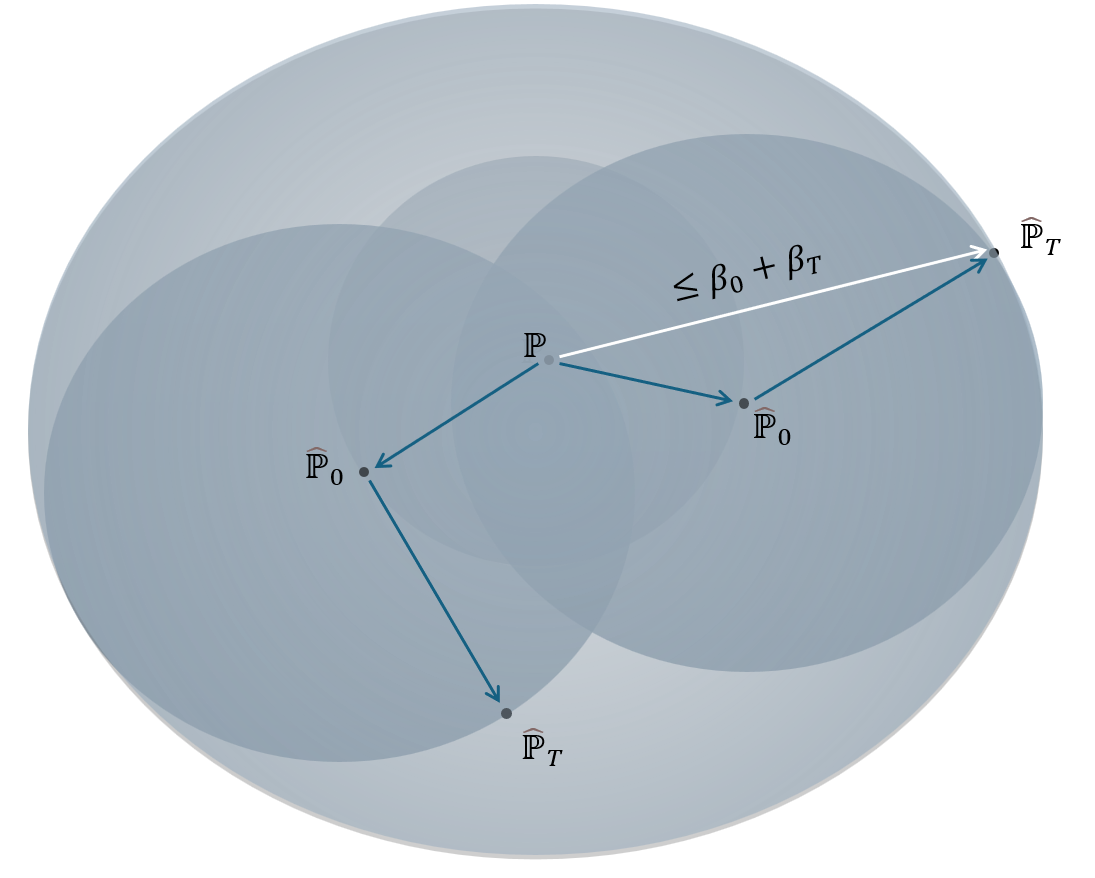}
\end{minipage}
        \caption{\textbf{Left:} Illustration of two possible samples $\hat{\mathbb P}_T$ after $T$ iterations of reciprocal learning. \textbf{Right:} The Wasserstein ball on the sample distortion in reciprocal learning. This ball bounds the distance between the sample $\hat{\mathbb P}_{T}$ at iteration $T$ and the law $\mathbb P$ by $\beta_0 + \beta_T$ (as illustrated in Figure~\ref{fig:illu-wasserstein-space}) via the triangle inequality. %
        }
        \label{fig:illu-bounds}
\end{figure}

Lemma~\ref{lemma:recipr-distortion} provides us with a bound on how far the learning algorithm moves the initial \iid~sample $\hat{\mathbb P}_0$ in Wasserstein space for any iteration $T$. We thereby establish a bound $\beta_T$ on the Wasserstein distance $W_p(\hat{\mathbb P}_0, \hat{\mathbb P}_{T})$ that is valid for any iteration $T$. It then follows directly that we can bound the Wasserstein distance between $\mathbb P$ and $\hat{\mathbb P}_{T}$ by $\beta_0 + \beta_T$ via the triangle inequality (see Figure~\ref{fig:illu-bounds}). The rest of our strategy is to relate the Wasserstein distance between distributions to the distance between corresponding risks via the Kantorovich--Rubinstein lemma \citep{kantorovich1958space}. Detailed reasoning about this argument is given in Section~\ref{sec:main}.

An apparent limitation of the triangle inequality argument in our strategy is that it does not capture the possibility that the movements in Wasserstein space (from $\mathbb P$ to $\hat{\mathbb P}_0$ and from $\hat{\mathbb P}_0$ to $\hat{\mathbb P}_T$) might cancel each other out. This is the best case from a generalization perspective: We might hope that the algorithm's sample adaptation would correct the finite-sample error in $\hat{\mathbb P}_0$ and push the sample towards $\mathbb P$. Ensuring this, however, requires assumptions on the distribution of the adaptively selected data. Such assumptions %
cannot generally be verified with certainty from finite observations. Practitioners might \textit{hope} that the query function has full access to $\mathbb P$ (as in active learning) or that the estimated pseudo-labels (as in semi-supervised learning) are correct, but we simply do not and will not know. 

Instead of relying on such presumptive assumptions, we thus base our analysis on \textit{verifiable conditions} (Conditions~\ref{ass-cont-diff-feat-param}--\ref{cond:lipschitz-sample-adpat} below) concerning something we completely control: the learning algorithms themselves. A practitioner can confirm these conditions hold for the algorithm they are using. Thus, we anchor our analysis in the initial \iid~sample~$\hat{\mathbb P}_0$ and refrain from any additional distributional assumptions on $\hat{\mathbb P}_t$ for $t > 0$. All we require are conditions on how the algorithms change $\hat{\mathbb P}_t$ to $\hat{\mathbb P}_{t+1}$. Some of these conditions, arguably, might be perceived as strong. Note, however, that they restrict specific algorithms rather than \textit{classes of} algorithms. That is to say, we do not exclude %
any of the particular examples that fall under the reciprocal learning framework (like boosting, bandits, active or semi-supervised learning), but only specific instantiations of these examples.
Thus, our strategy allows us to establish generalization bounds that are universal in two ways: %
They neither restrict the class of learning methods under consideration, nor make any assumptions on the distribution of newly added data at iterations $t \ge 1$.

Nevertheless, to show the generality and utility of our analysis strategy, as well as to demonstrate the potential for future work which builds on this paper, we provide an example (Proposition~\ref{prop:boosting}) illustrating how tighter bounds can be obtained with additional assumptions. These results require structure on the added data that arises in the context of boosting to provide stronger results on the generalization gap and excess risk.
Along the way to our results, our analysis embeds the reciprocal learning framework into statistical learning theory via multi-shot ERM (Section~\ref{sec:recirp-learning-theory}). This might be of independent interest, since it provides a novel bridge between feedback-driven online algorithms like active learning or semi-supervised learning on the one hand and learning theory on the other. %

Beyond this theoretical motivation, there is a strong practical rationale for generalization bounds on learning from adaptively selected samples. This rationale is threefold. First and foremost, our generalization bounds hold for a very broad class of machine learning algorithms, comprising active learning, semi-supervised learning, bandit algorithms, Bayesian optimization, boosting and superset learning (also known as \textit{partial label learning}). \citep[We refer the reader to][for an extensive list of algorithms which fall within the framework of reciprocal learning.]{rodemann24reciprocal} Naturally, they are not tight for all of these algorithms, but our general perspective answers the open question of whether generalization is even possible under data--parameter feedback: If we give algorithms enough degrees of freedom to alter the sample they are learning from, can they still generalize well beyond that adaptively selected sample? Our theorems offer first insights into this question. Crucially, these results have valuable implications for the practical design of novel reciprocal learning methods, on top of the insights they provide regarding existing methods, such as active learning or boosting (see Section~\ref{sec:conclusion} for details). 

Secondly, our results provide \textit{robust} generalization analyses for reciprocal learning methods. Consider active learning as an example. If the oracle provides true labels, our generalization bounds will not be tight. If, however, there is reason not to trust the oracle, a robust \citep[in the classical sense of][]{huber1981robust} analysis is required. Such an analysis will need to take into account potential contaminations of the oracle's labels. In this case, our generalization bounds prove useful as they hold for any kind of sample distortion, whether based on true labels, contaminated labels from an untrustworthy oracle, or otherwise. %

Thirdly, our results give rise to stopping criteria for all instances of reciprocal learning. In addition to bounds on convergent solutions, we give anytime-valid bounds. Since these latter bounds are iteration dependent, they can tell practitioners  when to stop a learning algorithm in order to remain below some generalization error with high probability. Such stopping criteria are essential to any safety-critical application---for example, semi-supervised learning for pipeline failure detection \citep{alobaidi2022semi} (see Section~\ref{sec:example}), active learning to identify patient safety events \citep{fong2017using}, or high-alert drug screening by boosting classifiers \citep{wongyikul2021high}, to name only a few.

The remainder of this article is structured as follows. After a review of related literature in Section~\ref{sec:related-work}, we introduce the main concepts required for our analysis in Section~\ref{sec:recirp-learning-theory}: covering entropy integrals and the technical definition of the reciprocal learning framework. These concepts allow us to embed this framework into statistical learning theory as a variant of multi-shot ERM. Section~\ref{sec:main} begins by introducing the conditions required for our results, and then answers the central question of how well a learner that adaptively selects data can generalize by establishing generalization bounds on convergent solutions (Section~\ref{sec:bounds-convergent-sol}) as well as anytime-valid generalization bounds (Section~\ref{sec:bounds-anytime}). 
The subsequent Section~\ref{sec:example} exemplifies these bounds for the special case of semi-supervised learning. We conclude with a brief discussion of our results and pointers to future work in Section~\ref{sec:conclusion}. All proofs are relegated to Appendix~\ref{app:proofs}.

\section{Related Work}\label{sec:related-work}

The questions of whether, and to what degree, inference from  adaptively selected samples can be valid have long occupied scientists. Pioneers of experimental design \citep{fisher1966design} and sampling theory \citep{cochran1942sampling} have recognized that in a myriad of scientific inquiries, the researcher can actively \citep[in the case of][]{fisher1966design} or inadvertently \citep[in the case of][]{cochran1942sampling} select the sample from which they draw conclusions. (See also Appendix~\ref{app-history} which provides a brief historical background on feedback-driven learning in the cybernetics community.) When the (self-)selection probabilities $\pi_i$ of units $x_i$ in a sample $x_1, \dots, x_n$ are known, \citet{horvitz1952generalization} proved the unbiasedness of the mean estimator
$\frac{1}{n}\sum_{i = 1}^n \pi_i^{-1} x_i $ (which was later named the Horvitz--Thompson estimator). 
\citet{heckman1979sample} introduced a more general two-step M-estimation correction for biased samples.

The statistics community has built on this work, giving rise to a huge strand of literature on selection bias, particularly in the context of survey statistics \citep[see][for a collection of such works]{wainer2013drawing}. Here, selection bias is the norm rather than the exception \citep{yerushalmy1972self}, since it arises in any situation \say{when a rule other than simple random sampling is used to sample the underlying population that is the object of interest. \ldots Distorting selection rules may be the outcome of decisions of sample survey statisticians, self-selection decisions by the agents being studied or both} \cite[page 201]{heckman1979sample}. Statistical inference from data collected by adaptive experimentation has been studied by \citet{zhang2021statistical,zhang2023statistical} with a focus on general $M$-estimation, and by \cite{nie2018adaptively} with a more specific focus on treatment effect estimation. 

The article at hand addresses a third source of distorting selection rules: adaptive \emph{algorithmic selection} decisions by machine learning algorithms like semi-supervised learning, bandits, boosting, active learning and others.
Many remedies for inference under selection bias, however, work regardless of the source of the bias, rendering the statistical literature on selection bias closely related to this article. For instance, \cite{chauvet-15} proved a central limit theorem for the Horvitz--Thompson estimator in a multistage sampling scenario similar to the framework of reciprocal learning: Simple random sampling without replacement is conducted in the first stage of sampling while later stages can have arbitrary sampling designs. \cite{chauvet-15} linked this design to samples where the primary sampling units are selected independently via coupling methods. 
\cite{han-wellner-21} gave Glivenko--Cantelli and Donsker theorems for Horvitz--Thompson empirical processes. Our analysis is more specific than both these works, as reciprocal learning algorithms provide us with more information on how exactly the sampling strategy changes after initial random sampling.

By and large, the machine learning community has mainly focused on how to learn parameters from data. Recently, however, there has been an increasing interest in the other side of the coin: How to efficiently (sub-)select training data---often in light of previously learned parameters. Concerns about degrading data quality and quantity for large models \citep{muennighoff2024scaling,mauri2021estimating} have inspired work on subsampling \citep{lang2022training,stolz2023outlier}, coresets \citep{pooladzandi2022adaptive,xiong2024fair,rodemann2022not}, data subset selection \citep{yin2024embrace,rodemann2022levelwise,chhabra2023data}, and data pruning \citep{zhang2024heprune,he2023candidate,ben2024distilling}.
While these approaches demonstrate remarkable experimental success, little is known about their theoretical guarantees---with the notable exception of \cite{durga2021training} for the special case of data subset selection. The article at hand has a broader scope. By proving generalization bounds for a wide range of data-selecting algorithms like active learning or self-training, this article provides first general insights into how well one can generalize from a model trained on data whose selection has been informed by the very same model.

Situated at the intersection of statistics and machine learning, several scholars have studied what we called adaptive {algorithmic selection} above. \cite{fithian2014local} proposed local case-control sampling techniques to select informative samples from imbalanced datasets, guided by preliminary model fits. Valid inference is then obtained via post-hoc analytic adjustment to the parameters. %
\citet{sarndal2003model} and \citet{breidt-17} review model-assisted inference from complex samples, including machine learning models. \cite{toth2011building} established asymptotic design $L_2$ consistency for regression trees estimated from complex samples, while \cite{nalenz2024learning} proposed debiasing methods for both trees and random forests under complex samples. 
For reciprocal learning's special case of semi-supervised learning, inferential guarantees have been studied by \cite{pepe1992inference,wasserman2007statistical,NIPS-unlabeled,zhang2019semi, song2024general}. Recently, \citet{angelopoulos2023prediction,angelopoulos2023ppi++} and \citet{zrnic2024cross} extended these results to the non-asymptotic case and to distribution shift under their (more general) prediction-powered inference framework.

\subsection{Connections to Specific Theory for Reciprocal Learning's Special Cases}
Beyond the literature on semi-supervised learning outlined immediately above, learning theory for multiple special cases of reciprocal learning has been developed in the past decades. Our contribution is complementary to, rather than a replacement for, paradigm-specific
results which analyze reciprocal learning's special cases. 

We will now briefly review results for active learning, boosting, bandits, superset learning, self-training and Bayesian optimization. 
For {active learning}, sharper guarantees are known under disagreement- and noise-based assumptions \citep{hanneke2015minimax} which our results do not require. For {boosting}, there are
classical and refined margin-based generalization theories \citep{schapire1998boosting,gao2013doubt}.
For (contextual) {bandits}, guarantees typically rely on propensities, overlap,
and off-policy evaluation arguments \citep{swaminathan2015batch,wang2017optimal}. These are, again, assumptions which we do not require.
For {superset learning}, consistency and estimation-error results are
available under explicit candidate-label generation models \citep{cour2011partial,liu2014superset,feng2020provably}.
For {self-training}, prior theory exists as well, including structural analyses and
results for specific pseudo-labeling schemes \citep{wei2021selftraining,he2023pseudolabeling}. %
Yet \citet{wei2021selftraining} do not give generalization bounds, while the generalization bounds in \citet{he2023pseudolabeling} hold only in expectation and rely on the true population law $\mathbb P$, which means they are typically not computable.
And for {Bayesian optimization}, the dominant theory is largely regret- and
convergence-based \citep{srinivas2010gpucb,berkenkamp2019noregret}, unrelated to generalizing from the data that is adaptively collected by Bayesian optimization, which is what our bounds address.

We do not aim to improve upon these results when their stronger structural
assumptions are justified, but give alternatives when they are not. We ask whether one can derive high-probability
generalization-gap and excess-risk guarantees {across} such sample-adapting learning
protocols \textit{without} assumptions on
the distribution or quality of newly added, reweighted, pseudo-labeled, or disambiguated
data. In this sense, our bounds are best viewed as assumption-lean baseline guarantees.
Their novelty is strongest for Bayesian optimization, but still substantial for self-training (see related work discussed above). For other methods, our work's usefulness primarily lies in giving almost-assumption-free, and thus robust, alternatives to existing bounds, as well as in providing a common analysis of inference from an
adaptively modified sample. Specifically, our bounds can be seen as robust alternatives to existing generalization bounds through the following lens. Intuitively, special cases of reciprocal learning imply special assumptions. These additional assumptions might then allow for tighter generalization bounds as in the example of active learning mentioned above. However, these assumptions typically refer to the newly added or altered data---e.g., the label annotation process (active learning) or the reward function (bandits), to name
two commonly required conditions. As these are distributional assumptions on population quantities (i.e., not fully observable), they cannot be verified with certainty and might be violated in practice. With regard to the altered data, our results only require verifiable assumptions on something we have complete control over: the adaptive algorithm itself. %

\subsection{Connections to Online Learning}
From an algorithmic perspective, our setting is connected to online learning, since both study sequential
interaction. Classical online learning, however, is usually analyzed through
\emph{regret} relative to a comparator under a stream of instances revealed round by round
\citep{cesa2004generalization,shalev2012online,neu2010online}. A complementary line of work shows how
online guarantees can be converted into statistical ones via online-to-batch or online-to-PAC
arguments \citep{neu2015explore,cutkosky2019anytime,lugosi2023online,abeles2025graph}. Our focus is
different in two respects. First, we do not study prediction on a stream of exogenous examples,
but repeated ERM on an \emph{initially i.i.d.} sample that is subsequently modified by the
algorithm through an endogenous (and fully observable) sample adaptation map. Second, our target is not a regret bound, nor a
conversion from regret to batch risk, but direct bounds on the generalization gap and excess
risk with respect to the \emph{original} population law $\mathbb P$ after the training sample has been
altered.

\subsection{Connections to Reinforcement Learning}
Reinforcement learning is another neighboring field which involves sequential feedback between
a learner and future data, typically formalized through the agent--environment interface of a
Markov decision process (MDP) \citep{sutton2018reinforcement}. The closest connections with our setting
are contextual bandits and related off-policy evaluation problems, where data are collected
adaptively and later used for statistical evaluation \citep{uehara2022review}. (See also regularized MDPs, \citealp{neu2017unified}.) Nevertheless, there is still a key distinction with the types of methods we are interested in: in full MDP-based reinforcement learning, the feedback loop is mediated by environment
dynamics, whereas in our framework the primary
object that changes is the \emph{sample} itself. Our results therefore address a different
inferential question: how to generalize from an adaptively modified training sample back to the
original population law, without assuming a model for the environment dynamics or for the
distribution of newly added data. In self-training, boosting, bandits and other reciprocal learning methods, only the \textit{training} data (sample) changes in response to the model. In reinforcement learning, the sample also changes, but simply because it is drawn from the new (changed) environment. 
This change in environment typically leads to a change in the target of inference: one is interested in how the learner performs under the distribution induced by the current policy rather than under the law that generated the initial sample. (Off-policy evaluation of reinforcement learning considers a fixed target policy, meaning that the target does not change. Nevertheless it is still not concerned with performance under the law that generated the initial sample.)
The motivation for the present work, however, is to give bounds on generalizing from some shifted empirical $\hat{\mathbb P}_t$ to the initial (unchanged) law $\mathbb P$.

\subsection{Connections to Out-of-Distribution (OOD) Generalization}\label{sec:OOD}

Another related body of work studies OOD generalization \citep{liuOutOfDistributionGeneralizationSurvey2023}. In this setting, there is distribution shift: a learner is trained on data drawn from a ``source'' distribution but is evaluated on data from a different, ``target'' distribution \citep{quinonero-candelaDatasetShiftMachine2022}. %
The literature on OOD generalization studies how well machine learning can perform in this setting, often as a function of the type and extent of the distribution shift. This includes the problem of domain generalization, where the goal is to design an algorithm that can be trained on one or more source ``domains'' (distributions) and then performs well on a new, unseen target domain \citep{wangGeneralizingUnseenDomains2023, singh2024domain, blanchardDomainGeneralizationMarginal2021, pmlr-v28-muandet13, zhouDomainGeneralizationSurvey2023}. A related field is domain adaptation, in which there is some data from the target domain available---either unlabeled data, a small amount of labeled data, or a combination of both \citep{wilsonSurveyUnsupervisedDeep2020, ben-davidTheoryLearningDifferent2010, zhaoReviewSingleSourceDeep2022, pmlr-v28-zhang13d}. In all cases, the object of interest is the new, shifted population. In contrast, we study generalization to the original, source population. Another difference between our setup and the bulk of the distribution shift literature is that the latter typically considers the shift in (test-time) distribution to be \textit{exogenous}, while reciprocal learning is concerned with \textit{endogenous} shifts of the training sample---namely, by the reciprocal learning algorithms themselves. Moreover, OOD generalization does not explicitly consider that the training process is iterative, a property which is central to our work.

The setting studied in the present work does not allow for changes in the underlying population distribution, as is the case for OOD generalization, domain generalization and domain adaptation. Instead, we consider learning procedures in which the training sample itself is adaptively modified during the learning process, for example by selecting or generating additional observations based on previously learned parameters. That is to say, the distinguishing feature of our setup is that the training sample is being modified, while the distinguishing feature of the distribution-shift setup is that there is a mismatch between source and target distributions.
This difference results in substantively different types of generalization bounds. Nevertheless, there are conceptual connections, since some methods developed for distribution shift problems implicitly modify the effective training distribution through mechanisms such as data augmentation and sample/domain reweighting. From this perspective, such procedures may be interpreted as instances of learning from adaptively selected data, suggesting that the framework studied here may also provide insights for certain forms of algorithmically induced distribution shift.

\section{Reciprocal Learning Theory}\label{sec:recirp-learning-theory}

We start by embedding reciprocal learning into statistical learning theory \citep{vapnik1998statistical,vapnik1968uniform}, which will allow us to reason about how well reciprocal learning generalizes. 
Specifically, we postulate a true but unknown data generating distribution $\mathbb P$ on the space $\mathcal Z = \mathcal X \times \mathcal Y$, where $\mathcal X \subset \mathbb R^{\dx}$ and $\mathcal Y \subset \mathbb R^{\dy}$. %
We define $\dz := \dx + \dy$ %
and we assume that $\mathcal Z$ is bounded (with respect to the Euclidean norm $\| \cdot\|_2$). %
As is customary, we denote labels by $y \in \mathcal Y$ and feature vectors by $x \in \mathcal X$, while random variables on these spaces are given by $Y$ and $X$, respectively.
We consider a set of measurable functions 
\[\mathcal F := \{ f_\theta : \mathcal X \to \mathcal Y \mid \theta \in \Theta\},\]
parametrized by a subset $\Theta$ of Euclidean space. %
(Note that $\mathcal F$ is uniformly bounded---i.e., $\sup_{\theta \in \Theta, x \in \mathcal{X}} \allowbreak \|f_\theta(x)\|_2 < \infty$---because $\mathcal Y$ is bounded by assumption.) %
Following standard terminology, we refer to $\mathcal{F}$ as the \textit{hypothesis space} and to $\Theta$ as the \textit{parameter space}. %
We call $f_\theta$ a \emph{prediction function} or \emph{predictor}, and $\theta$ a \emph{parameter vector}. We denote the set of Borel probability measures on $\mathcal Z$ by $\mathcal P(\mathcal Z)$. We define $\mathcal P(\mathcal X)$ and $\mathcal P(\mathcal Y)$ similarly. The true data generating distribution $\mathbb P$ is taken to be an element of $\mathcal P(\mathcal Z)$.

Statistical learning theory requires a choice of a complexity measure to describe the richness of $\mathcal{F}$. In this work, we use the covering entropy integral %
\citep{kolmogorov1959varepsilon,talagrand2014upper}.

\begin{definition}[Covering Entropy Integral]\label{def:covering-n}
Let $\|\cdot\|$ %
denote a norm on $\mathcal F$, such as the essential-supremum norm
\[\left\|f\right\|_{L_\infty(P)}:=\operatorname{\textit{P}-ess\,sup}_{x \in X} \|f(x)\|_2,\] 
or the $L_2$ norm 
\[\|f\|_{L_2(P)} := \sqrt{\mathbb E_{X \sim P} [\|f(X)\|_2^2]},\] 
with respect to some measure $P \in \mathcal{P}(\mathcal X)$. %
For $\epsilon>0$, define
the \emph{covering number} $\mathcal{N}(\mathcal{F},\|\cdot\|, \epsilon)$ %
as the smallest $N \in \mathbb N$ such that there exists 
$\theta_1, \ldots, \theta_N \in \Theta$ 
satisfying %
$\min _{1 \leq k \leq N}\left\|f_\theta-f_{\theta_k}\right\| \leq \epsilon$ for all $\theta \in \Theta$. 
The \emph{uniform} and \emph{$L_2$ covering entropy integrals} are then respectively defined as
\begin{align*}
    \mathfrak{C}_{L_\infty(P)}(\mathcal{F}) &:= \int_{0}^{\infty} \sqrt{\log \mathcal{N}\left(\mathcal{F},\|\cdot\|_{L_\infty(P)}, \varepsilon\right)} \mathrm{d} \varepsilon, \\
    \mathfrak{C}_{L_2}(\mathcal{F}) &:= \sup_{P \in \mathcal{P}(\mathcal X)} \int_{0}^{\infty} \sqrt{\log \mathcal{N}\left(\mathcal{F},\|\cdot\|_{L_2(P)}, \varepsilon\right)} \mathrm{d} \varepsilon.
\end{align*}
\end{definition}
One interpretation of the covering number $\mathcal{N}(\mathcal{F},\|\cdot\|, \epsilon)$ is as the minimum number $N$ of balls $
    B\left(f_{\theta_k}, \epsilon\right)=\left\{f_\theta \in \mathcal{F}:\left\|f_\theta - f_{\theta_k}\right\| \leq \epsilon\right\} 
$ with radius $\varepsilon$ %
 required to cover $\mathcal{F}$ (in the sense that $
    \mathcal{F} \subset \bigcup_{k=1}^N B\left(f_{\theta_k}, \epsilon\right)
$).
The radius $\varepsilon$ can be considered as the resolution, or scale of the covering number. By integrating over $\varepsilon$, the covering entropy integral collects information over every scale. Using the Lebesgue measure on $\varepsilon$ (i.e., integrating uniformly) means that we are not privileging one scale over another. %
Although the uniform covering entropy integral $\mathfrak{C}_{L_\infty(P)}(\mathcal{F})$ is more common, we use the \(L_2\)-covering entropy integral $\mathfrak{C}_{L_2}(\mathcal{F})$ because, evidently, $\mathfrak{C}_{L_2}(\mathcal{F})$ does not depend on $\mathbb P$. %
This makes it particularly appealing for our endeavor, since the learning methods we consider can operate outside of the support of the true $\mathbb P$ by changing the sample. This is also the main reason why we prefer it over more popular complexity measures like Rademacher complexity \citep{bartlett2002rademacher, bartlett2005local}. Further note that the Rademacher complexity can be upper-bounded by the covering entropy integral---we will in fact use this property in proving Theorem~\ref{thm:excess-risk-classic} later.

Before turning to reciprocal learning in detail, we still have to define three pivotal functions---the learner, the loss and the risk---that allow one to find and assess a parameter $\theta$ and its corresponding prediction function $f_\theta \in \mathcal{F}$ with respect to any law $P \in \mathcal{P}(\mathcal Z)$. (Contrast this $P$ with the fixed true law $\mathbb{P}$.)

\begin{definition}[Learner]
    A \emph{learner} is a function $\hat \theta : \bigcup_{n \ge 1} \mathcal Z^n \rightarrow \Theta$, 
    which maps samples $(Z_1, \ldots, Z_n)$ of size $n$
    to parameters $\hat \theta(Z_1, \ldots, Z_n) \in \Theta$. (Note that, to avoid excessive notation, we sometimes denote the fitted parameter $\hat \theta(Z_1, \ldots, Z_n)$ by $\hat \theta$ when the context ensures that there would be no confusion between it and the learner.)
\end{definition}
A central ingredient of statistical learning theory is the \emph{loss function}. In our setup, a
loss function is a measurable map
$\ell:\Theta \times \mathcal Z \to \mathbb R$, %
where $\ell(\theta, x,y)$ quantifies the cost incurred by the parameter $\theta\in\Theta$ at an
observation $(x,y)\in\mathcal Z$. This setup is slightly more general
than the standard prediction-loss formulation---where the loss $\tilde{\ell}( f_\theta(x),y)$ is computed by directly comparing the prediction $f_\theta(x)$ with the true label $y$. 
Clearly, the standard formulation $\tilde{\ell}$ is included as a special case of our setup, since we can define $\ell(\theta,x,y)=\tilde{\ell}\bigl(f_\theta(x), y\bigr)$.
Common examples covered in this way include the squared loss in regression, logistic (cross-entropy) loss in probabilistic classification, and hinge loss in margin-based classification. Our setup also allows for regularization through a penalized loss function.

\begin{definition}[Risk and Loss]\label{def:loss}
The \emph{risk} of a parameter $ \theta \in \Theta$ on data generated according to a probability $ P \in \mathcal{P}(\mathcal Z)$ is defined as
\[
\mathscr R(P, \theta) :=  %
\int_{\mathcal Z} \ell(\theta, x, y) \mathrm{d} P(x,y),
\] 
where $\ell : \Theta \times \mathcal Z \rightarrow \mathbb R$ is the \emph{loss function}. %
\end{definition}

The loss function $\ell$ induces a function class 
\begin{equation}\label{eq:defnCenteredLossFunctionClass}
    \mathcal L_0 : = \{ z \mapsto \ell (\theta, z) - \ell(\theta, z_0) : \theta \in \Theta\},
\end{equation}
which is the class of centered loss functions parametrized by $\theta \in \Theta$ and centered at some fixed $z_0 \in \mathcal Z$. The covering entropy integral $\mathfrak C_{L_2}(\mathcal L_0)$ will be central to our excess risk bounds, as we will use it to bound the difference between the risk on the sample and the true risk (on $\mathbb P$). %

Given training data $Z_{1}, \ldots, Z_{n} \iidsim \mathbb P$, standard machine learning revolves around finding a learner which outputs a parameter estimate $\theta \in \Theta$ %
whose risk $\mathscr R(\mathbb P, \theta)$ is close to the unknown minimum risk, often referred to as the \emph{Bayes risk}\footnote{Not to be confused with the concept of Bayes risk in statistical decision theory, which is not the minimum risk but the risk integrated over a prior $\pi$, i.e., $B(\pi, \theta) := \mathbb E_{\mathbb P \sim \pi} [ \mathscr R(\mathbb P, \theta)]$.%
}
\begin{equation*}%
\mathscr R_\Theta:=\inf _{\theta \in \Theta} \mathscr R(\mathbb P, \theta).
\end{equation*}
The difference $\mathscr R(\mathbb P, \theta) - \mathscr R_\Theta$ is called the \textit{excess risk} of $\theta$. The quantity $\mathscr R(\mathbb P, \theta) $ is commonly referred to as the \emph{generalization error} of the parameter $\theta$ \citep{shalev2014understanding}. 
A popular learner in the literature is one-shot ERM:

\begin{definition}[One-Shot ERM]\label{def:ERM}
Denote by $\hat{\mathbb P}:=\frac{1}{n} \sum_{i=1}^{n} \delta_{Z_{i}}$ %
the \emph{empirical distribution} of the training data $Z_1, \ldots, Z_n$, where $\delta_z$ is the Dirac measure at $z \in \mathcal Z$. A learner $\hat \theta$ is an \emph{empirical risk minimizer} if 
\[\hat{\theta}(Z_1, \dots, Z_n)  \in \argmin_{\theta \in \Theta}
\mathscr R(\hat{\mathbb P}, \theta)= \argmin_{\theta \in \Theta} \frac{1}{n} \sum_{i=1}^{n} \ell( \theta, x_i, y_i).\]
\end{definition}
Through the lens of Definition~\ref{def:ERM}, reciprocal learning can be understood as a specific variant of \textit{multi}-shot ERM, as detailed in the following definition: %

\begin{definition}[Reciprocal Learning and Sample Adaptation]\label{def:basic-reciprocal-learning}
A \emph{sample adaptation function} $f_s : \Theta \times \mathcal{P}(\mathcal Z) \to \mathcal{P} (\mathcal Z)$ %
is a (possibly stochastic) map from the current fitted model and empirical distribution (i.e., sample) to a new empirical distribution. A \emph{reciprocal learning algorithm} is a process $(\hat \theta_0, \hat{\mathbb P}_0), \ldots, (\hat \theta_T, \hat{\mathbb P}_T)$ defined by its sample adaptation function $f_s$, where 
\begin{enumerate}
    \item $\hat{\mathbb P}_0$ is the empirical distribution of some initial training data $Z_1, \ldots, Z_n \iidsim \mathbb P$ with sample size $n$;%
    \item\label{defn:ERM-step} For $t=0,\ldots,T$, the fitted parameter $\hat \theta_{t}$ is an empirical risk minimizer---that is, $\hat \theta_t \in \argmin_{\theta \in \Theta} \mathscr R(\hat{\mathbb P}_{t}, \theta)$ (the ERM step); and
    \item For $t=1,\ldots,T$, the empirical distribution $\hat{\mathbb P}_{t}$ is given by $f_s(\hat \theta_{t-1}, \hat{\mathbb P}_{t-1})$ (the sample adaptation step).%
\end{enumerate}
Let $n_t$ be the sample size at iteration $t$ (so that $n = n_0$) and $m_t$ be the number of sample points which are changed at iteration $t \ge 1$ (i.e., in the transition from \(\hat{\mathbb P}_{t-1}\) to \(\hat{\mathbb P}_t\)):
\[m_t := \sup_A \left| n_t \hat{\mathbb P}_t(A) - n_{t-1} \hat{\mathbb P}_{t-1}(A) \right|,\]
where the supremum is over all Borel $A \subset \mathcal Z$. (Clearly $m_t  \leq \max\{n_{t-1},n_t\}$.)
We write $m:=m_1$ for the number of points changed by the first sample adaptation.\footnote{$m_1$ is the only per-iteration change our results depend on, since transitions at $t\ge 2$ are controlled by Condition~\ref{cond:lipschitz-sample-adpat}.}%

\end{definition}
That is to say, reciprocal learning performs $T+1$ ERMs, producing fitted parameters $\hat \theta_t$, $t \in \{0, \dots, T\}$, in which the empirical distribution $\hat{\mathbb P}_{t}$ at time $t \ge 1$ is affected by both $\hat{\mathbb P}_{t-1}$ and $\hat \theta_{t-1}$ through sample adaptation. %

We do not assume there is a unique empirical risk minimizer in Step~\ref{defn:ERM-step} of Definition~\ref{def:basic-reciprocal-learning}, just that one (or more) exists at each iteration of reciprocal learning. However, some of our results require strong convexity (Condition~\ref{ass-str-convex} below) which implies the empirical risk minimizer exists and is unique when $\Theta$ is closed and convex. %
(Whenever the empirical risk minimizer is nonunique, we understand the reciprocal learning algorithm to include a fixed rule for selecting $\hat \theta_t$ from the multiple empirical risk minimizers.)

\begin{figure}
    \centering
    \includegraphics[width=0.9\linewidth]{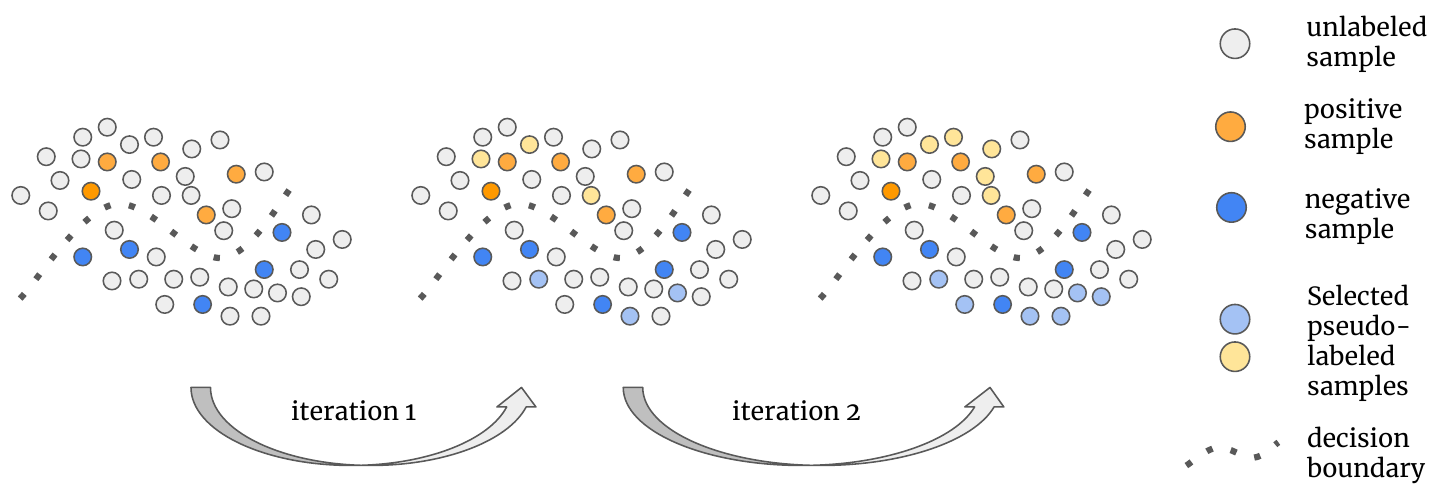}
    \caption{A simple example of reciprocal learning: Self-training for binary classification. Features $x$ are chosen based on current model $\theta$, then added to the sample together with self-predicted (pseudo-)labels $\hat y(\theta,x)$. (Adapted with permission from \citealp{goschenhofer2023reducing}.)}%
    \label{fig:self-training}
\end{figure}

For the sake of illustration, consider the paradigm %
of self-training in semi-supervised learning. As described in Section~\ref{sec:intro}, self-training algorithms iteratively add pseudo-labeled data to the training sample, as follows: 1) Fit a model to the fully labeled training data. 2) Use this model to predict labels for the unlabeled training data. 3) Select a subset of these unlabeled data. 4) Add this subset together with the predicted labels (so-called pseudo-labels) to the sample and refit. 5) Iterate Steps 2--4 using the refitted model. Figure~\ref{fig:self-training} illustrates this process in the simple case of binary classification. Evidently, the sample at iteration $t+1$ of this process depends on both the sample and the model at iteration $t$, and so %
self-training is clearly an instance of reciprocal learning. To see this dependency more explicitly, consider self-training's sample adaptation function. In the special case where a single sample point is added at each iteration (so that $m_t = 1$), at iteration $t + 1$ this sample adaptation function %
outputs the empirical distribution 
\[\hat{\mathbb P}_{t+1}(X = x, Y = y) =  \frac{ \mathds{1} \big\{ x = x_s(\hat{\theta}_{t}), y =  \hat y(\hat{\theta}_{t}, x_s(\hat{\theta}_{t})) \big\} +  n_{t}  \hat{\mathbb P}_{t}(X = x, Y = y) }{n_{t}+1}, 
\]
where $\mathds{1}$ is the indicator function.
Here, $x_s(\hat{\theta})$ \citep[Definition~2 of][]{rodemann24reciprocal} is the unlabeled data point with the highest confidence score \citep{arazo2020pseudo,rizve2020defense,li2020pseudo} according to $\hat{\theta}$, %
and $\hat y: \Theta \times \mathcal X \rightarrow \mathcal Y$ is the prediction function, which returns the predicted label of the selected $x_s(\hat{\theta})$ based on the learned $\hat{\theta}$. 

Another prominent ML method is active learning \citep{ma2016online,holzmuller2023framework,sloman2024bayesian}. 
This learning method is similar to self-training, except that the assignment of labels $y$ to selected features $x$ is done by an oracle with access to the \say{ground truth} rather than by the fitted model. 
Thus, modifying the sample adaptation function for self-training given above to query an oracle
will result in a sample adaptation function for active learning. This shows that active learning also falls within the framework of reciprocal learning. Other learning methods can also be embedded in this framework in the same way---i.e., by constructing appropriate sample adaptation functions. %
\citep[We refer the reader to Appendix~A of][for a list of the sample adaptation functions corresponding to these other learning methods.]{rodemann24reciprocal}

\citet{rodemann24reciprocal} distinguish between greedy and non-greedy sample adaptation functions.\footnote{The usage of the term ``greedy'' and ``non-greedy'' should not be confused with the way these terms are used in the optimization and reinforcement learning literature, where ``greedy'' refers to rules that choose points so as to maximize a given functional.} While the former only \textit{adds} data, the latter only \textit{replaces} data points. %
For greedy sample adaptation, $m_t = n_{t} - n_{t-1} \ge 0$ for all $t \in \{1, \dots, T\}$. In the non-greedy case, by comparison, $n_t = n_{t-1}$ even though $m_t \ge 0$.
It is possible, of course, that a sample adaptation function can be neither greedy nor non-greedy---for example, if it removes data points without replacing them, or if it both adds and modifies training data.

The Lipschitz constant $L_s$ of the sample adaptation function $f_s$ plays a pivotal role for proving both convergence and generalization of reciprocal learning. Roughly speaking, $L_s$ formalizes the maximum amount of change in the sample a previous change in the sample or fitted parameter can induce. The higher the value of $L_s$, the more sensitive $f_s$ is towards previous changes in the sample or fitted parameter. (In this work, we care particularly about the initial change of the sample.)

Often, sample adaptation consists of a \emph{data selection} function $\Theta \rightarrow \mathcal{X}$ and a \emph{label assignment} function $\Theta \times \mathcal{X} \rightarrow \mathcal{Y}$.  
\citet{rodemann24reciprocal} give sufficient conditions for this kind of sample adaptation to be Lipschitz continuous when it only changes a single data point per iteration (i.e., when $m_t = 1$). %
These conditions require that both the data selection and the label assignment functions are Lipschitz continuous. %
In this setting, Lipschitz continuity can be thought of as a type of regularization. 
Intuitively, regularizing the data selection function prevents changes to the sample covariates $x$ that are ``too abrupt.'' More generally, regularizing the sample adaptation function smooths the dependence between the current parameter iterate and the next training sample, so that nearby parameters induce nearby samples. Such data regularization is symmetric to classical statistical regularization (e.g., $L_1$- or $L_2$-regularization, see \citealp[Chapter~5]{hastie2009elements}, or \citealp[Chapter~13]{shalev2014understanding}): %
It smooths out the effect of the fitted parameters on data adaptation, thus preventing ``overfitting'' \textit{of the sample to the (fitted) parameter}, while statistical regularization smooths out the effect of the data on parameter selection, preventing overfitting \textit{of the parameter to the sample}. (We say ``preventing overfitting'' as shorthand for the statement that Lipschitz continuity prohibits parameter-sensitive sample adaptation---i.e., it prohibits small changes in the current parameter inducing large changes in the subsequent sample, which makes the learning process unstable.)

Some reciprocal learning algorithms---such as many self-training or boosting algorithms---adapt the sample in a deterministic way, while others---such as epsilon-greedy search \citep{kuleshov2014algorithms} or the popular Thompson sampling \citep{russo2016information}---adapt it in a stochastic way. When the sample adaptation is deterministic, there is a single \emph{reciprocal path} $(\hat \theta_0, \hat{\mathbb P}_0), \ldots, (\hat \theta_T, \hat{\mathbb P}_T)$ for each starting sample $\hat{\mathbb P}_0$. In the case of random sample adaptation, both the \emph{sample path} $\hat{\mathbb P}_0, \dots, \hat{\mathbb P}_T$ and the reciprocal path are stochastic processes. As such,  reciprocal learning comprises multiple possible paths.
The Lipschitz condition on $f_s$ is to be understood pathwise in this scenario; that is to say, $f_s$ should be almost surely Lipschitz continuous, with $L_s$ the essential supremum of the pathwise Lipschitz constants.

Technically, this Lipschitz condition on the sample adaptation function is with respect to the $p$-Wasserstein distance. This distance will be required in our generalization analysis later, so we define it here. %

\begin{definition}[$p$-Wasserstein Distance]\label{def:wasserstein}
    For $p \in [1,\infty)$, the $p$-Wasserstein distance between $P, Q \in \mathcal P(\mathcal Z)$ %
    is defined as
    \[
    W_{p}(P, Q):= \inf_{\pi \in \mathcal C} \left( \int_\mathcal{Z} \| z - z^{\prime}\|_2^p \mathrm{d} \pi(z,z') \right)^{1 / p}, 
    \]
    where the infimum is taken over the set $\mathcal C$ of all couplings of $P$ and $Q$. (A coupling of $P$ and $Q$ is a probability measure $c \in \mathcal{C}$ on the product space $\mathcal{Z} \times \mathcal{Z}$ whose marginals are $P$ and $Q$.)  
\end{definition}      
The following characterization of the 1-Wasserstein distance by \citet{kantorovich1958space} will be instrumental for relating the 1-Wasserstein distance between distributions to the difference in corresponding risks later.

\begin{lemma}[Kantorovich--Rubinstein Lemma]\label{lemma:kant-rubin}
    \textnormal{\citep{kantorovich1958space}}
    For all $K > 0$ and $P,Q \in \mathcal P(\mathcal{Z})$, 
    \[\displaystyle W_{1}(P, Q)={\frac{1}{K}}\sup_{\|f\|_{L}\leq K}\mathbb{E}_{Z \sim P }[f(Z)]-\mathbb{E}_{Z\sim Q }[f(Z)],\] 
    where the supremum is over all Lipschitz-continuous (with respect to $\| \cdot \|_2$ on $\mathcal Z$) functions $f : \mathcal Z \to \mathbb R$ with Lipschitz norm $\| f \|_L$ at most $K$.
\end{lemma}
\citet{rodemann24reciprocal} show that, under some conditions, the iterates $ (\hat \theta_{t}, \hat{\mathbb P}_{t})$ of reciprocal learning converge:  
\begin{theorem}[Convergence of Reciprocal Learning]\label{thm:convergence}
    \textnormal{\citep[Theorem 3 of][]{rodemann24reciprocal}} If a reciprocal learning algorithm uses a non-greedy sample adaptation function $f_s : \Theta \times \mathcal{P} (\mathcal Z) \to \mathcal{P} (\mathcal Z)$ that is Lipschitz with a sufficiently small Lipschitz constant, and the loss is strongly convex as well as continuously differentiable, then the iterates $(\hat \theta_{t}, \hat{\mathbb P}_{t})$ of reciprocal learning converge at a linear rate as $t \to \infty$. %
\end{theorem}
We denote the limit of the reciprocal learning process $(\hat \theta_0, \hat{\mathbb P}_0), (\hat \theta_1, \hat{\mathbb P}_1), \ldots,$ by $(\hat{\theta}_c, \hat{\mathbb P}_c)$. The limit $(\hat{\theta}_c, \hat{\mathbb P}_c)$ does not depend on the initial sample $\hat{\mathbb P}_0$; that is to say, $(\hat \theta_c, \hat{\mathbb P}_c)$ is the same (constant) limit regardless of the realized value of the initial sample. Hence, we say that reciprocal learning's convergence is ``pointwise in $\hat{\mathbb P}_0$.'' When $f_s$ is stochastic, the convergence is ``almost surely pathwise in $(\hat \theta_t, \hat{\mathbb P}_t)$.'' By this we mean that reciprocal learning converges to $(\hat{\theta}_c, \hat{\mathbb P}_c)$ on almost every reciprocal path. Here, and elsewhere in the article, convergence of $(\hat{\theta}_t, \hat{\mathbb P}_t)$ to $(\hat{\theta}_c, \hat{\mathbb P}_c)$ is understood to mean that both $W_p(\hat{\mathbb P}_t, \hat{\mathbb P}_c)$ and $ \|\hat \theta_t - \hat \theta_c \|_2$ go to zero almost surely as $t \to \infty$ (Condition~\ref{cond:conv-reciprocal-learning} below).

\section{Can Reciprocal Learning Generalize?}\label{sec:main}

In the rest of this article, we will investigate how well reciprocal learning generalizes. In particular, we ask three questions. Firstly, can we establish bounds on the generalization gap $\mathscr R(\mathbb P, \hat{\theta}_c) - \mathscr R(\hat{\mathbb P}_c, \hat{\theta}_c) $, i.e., the difference between the model's final training error and its generalization error? 
Secondly, can we bound its excess risk $\mathscr R(\mathbb P, \hat{\theta}_c) - \mathscr R_\Theta $, i.e., the difference between the model's generalization error and the best achievable generalization error?
Thirdly, can we obtain bounds on the generalization gap and excess risk that hold for $\hat{\theta}_T$ at any iteration $T$, rather than for the limiting parameter $\hat \theta_c$? 
Answering those questions will require conditions on both the reciprocal learning algorithm and the loss function being used. Each of our results only requires a subset of the following conditions. Strong convexity (Condition~\ref{ass-str-convex}), for instance, is only required by Theorems~\ref{thm:excess-risk-classic} and~\ref{thm:excess-risk-classic-anytime}.

\begin{condition}[Regularity of Loss] \label{ass-cont-diff-feat-param}
    \regularityConditionLabel{ass-cont-diff-feat-param-Lips-data}\textbf{(A)} The loss function $\ell$ is $L_{\ell}$-Lipschitz continuous in the data---that is, $\|\ell(\theta, x, y) - \ell(\theta, x', y') \|_2 \le L_{\ell} \|(x,y) - (x',y')\|_2$ for all $x, y, x', y'$ and $\theta$.
    \regularityConditionLabel{ass-cont-diff-feat-param-cts-diff-param}\textbf{(B)} $\ell$ is $L_{\ell}$-Lipschitz in the parameter $\theta$. Further, it is continuously differentiable with respect to $\theta$. That is, the gradient $\nabla_{\theta} \ell(\theta, x, y)$ exists and is $\kappa$-Lipschitz continuous in $(x,y) \in \mathcal Z$ for every $\theta \in \Theta$ (with respect to the Euclidean distance on the domain $\mathcal Z$ and codomain $\mathbb R^{\mathrm{dim}(\Theta)}$).
\end{condition}

\begin{condition}[Strong Convexity]\label{ass-str-convex}
The loss function $\ell$ is $\gamma$-strongly convex. That is, 
$
\ell(\theta',x,y) \geq \ell\left(\theta,x,y\right)+\nabla_{\theta} \ell\left(\theta,x,y\right)^{\top}\left(\theta'-\theta\right)+\frac{\gamma}{2}\left\|\theta'-\theta\right\|_{2}^{2},
$
for all $\theta, \theta^{\prime}, x$ and $y$. %
\end{condition}

\begin{condition}[Convergence of Reciprocal Learning]\label{cond:conv-reciprocal-learning}
    The iterates $(\hat \theta_{t}, \hat{\mathbb P}_{t})$ of reciprocal learning converge almost surely to $(\hat{\theta}_c, \hat{\mathbb P}_c)$. More specifically, $W_p(\hat{\mathbb P}_t, \hat{\mathbb P}_c) + \|\hat \theta_t - \hat \theta_c \|_2 \to 0$ almost surely pathwise as $t \to \infty$.  %
\end{condition}

\begin{condition}%
[Lipschitz-Continuous Sample Adaptation]\label{cond:lipschitz-sample-adpat}%
Let $\hat \theta(P)$ be the reciprocal learner's fitted parameter given the distribution $P \in \mathcal P(\mathcal Z)$. Then the map $P \mapsto f_s(\hat \theta(P), P)$ is almost surely $L_s$-Lipschitz continuous with respect to the $p$-Wasserstein distance.
\end{condition}
Recall that \cite{rodemann24reciprocal} give criteria 
for reciprocal learning to converge (Condition~\ref{cond:conv-reciprocal-learning}) and for the sample adaptation function to be $L_s$-Lipschitz continuous (Condition~\ref{cond:lipschitz-sample-adpat}), and show that these criteria are fulfilled by numerous algorithms. %
However, while these criteria are sufficient, they are not necessary for convergence and Lipschitz continuity. We therefore work directly with Conditions~\ref{cond:conv-reciprocal-learning} and~\ref{cond:lipschitz-sample-adpat}, rather than requiring the sufficient criteria used by \citeauthor{rodemann24reciprocal} to establish them.
Notably, these sufficient criteria include Conditions~\ref{ass-cont-diff-feat-param} and~\ref{ass-str-convex}, among other requirements. %

Conditions~\ref{ass-cont-diff-feat-param}--\ref{cond:lipschitz-sample-adpat} do not restrict the class of reciprocal learning paradigms. This implies our results hold for any reciprocal learning paradigm, ranging from, e.g., active learning, to bandits and semi-supervised learning. Further note that we do not need any assumptions on the distribution of the iteratively added data. Our conditions only apply to the design of reciprocal learning algorithms---something that is completely in the control of the machine learning practitioner (see Section~\ref{sec:conclusion}). 
Our results do not need any further conditions beyond those listed above.

Conceptually, almost all that our bounds require is a surprisingly simple reinterpretation of Wasserstein ambiguity sets.
To explain this, note that these bounds will mainly depend on the quality\footnote{Measured by the distance between the sample and the true law in Wasserstein space (see Lemma~\ref{lemma:wasserstein-conv}).} and the quantity $n$ of the initial sample $\hat{\mathbb P}_0$ drawn~from~$\mathbb P$ as well as the degree of the sample adaptation's potential distortion of $\hat{\mathbb P}_0$ in $\hat{\mathbb P}_{T}$. %
We measure the degree of this distortion by the $p$-Wasserstein distance between these distributions. In particular, we rely on the notion of ambiguity sets in Wasserstein space \citep{mohajerin2018data,shafieezadeh2015distributionally}:

\begin{definition}[Ambiguity Set]\label{def:amb-set}
     For any $P \in \mathcal P(\mathcal Z)$, define the ambiguity set $\mathcal{A}_{\rho}( P)$ as the $p$-Wasserstein ball of radius $\rho \geq 0$ centered at $P$:
$$
\mathcal{A}_{\rho}(P) := \left\{Q \in \mathcal{P}(\mathcal Z) : W_{p}(P, Q) \leq \rho\right\}.
$$
\end{definition}
We need a meaningful ambiguity set centered at initial $\hat{\mathbb P}_0$ that describes all samples reachable by reciprocal learning algorithms. As it turns out, we can upper-bound the Wasserstein distance between $\hat{\mathbb P}_0$ and any $\hat{\mathbb P}_{T}$ in the following way.

\begin{lemma}[Reciprocal Distortion Bound]\label{lemma:recipr-distortion}
        Let $\hat{\mathbb P}_{T}$ be the empirical distribution of training data at iteration $T$ in reciprocal learning and $\hat{\mathbb P}_0$ the initial one. If the sample adaptation function $f_s$ is almost surely $L_s$-Lipschitz (Condition~\ref{cond:lipschitz-sample-adpat}), then

    \begin{equation}\label{eq:recipr-distortion}
            W_p(\hat{\mathbb P}_0, \hat{\mathbb P}_{T}) \leq  S_T(L_s) \left(\frac{m}{n}\right)^{\frac{1}{p}} \diamZ =: \beta_T, %

    \end{equation}
        pointwise in $\hat{\mathbb P}_0$ and almost surely pathwise in $\hat{\mathbb P}_T$, where $\diamZ := \sup_{z, z' \in \mathcal Z} \| z - z \|_2 < \infty$ %
    is the diameter of $\mathcal{Z}$ and  %

    \begin{equation}\label{eq:defn-geo-sum}
             S_T(L_s) = \begin{cases}
        \frac{L_s^{T} - 1}{L_s - 1} & \mathrm{if\ } L_s \ne 1,\\
        T & \mathrm{if\ } L_s = 1.
    \end{cases}

    \end{equation}
\end{lemma}
Proofs of all our results can be found in Appendix~\ref{app:proofs}. The proof of Lemma~\ref{lemma:recipr-distortion} uses the strategy described by Figure~\ref{fig:illu-bounds} in Section~\ref{sec:outline}. The bound~\eqref{eq:recipr-distortion} holds \emph{pointwise in $\hat{\mathbb P}_0$}---i.e., it applies to every realization of the initial sample---rather than, for example, holding with high probability over the sampling of $\hat{\mathbb P}_0$ from $\mathbb P$. This is because the proof involves no probabilistic argument itself. All it does is compare consecutive elements of a sample path $\hat{\mathbb P}_0, \dots, \hat{\mathbb P}_T$ via the triangle inequality and Condition~\ref{cond:lipschitz-sample-adpat}. For deterministic sample adaptation, the randomness in $\hat{\mathbb P}_T$ and $\hat{\mathbb P}_0$ come from the same source, namely the initial random sampling, with $\hat{\mathbb P}_T$ a deterministic transformation of random $\hat{\mathbb P}_0$. For each possible value of $\hat{\mathbb P}_0$, $f_s$ induces a unique sample path and the bound in Lemma~\ref{lemma:recipr-distortion} holds along this path, regardless of what the realized value of $\hat{\mathbb P}_0$ turns out to be. In the case of random sample adaptation, the randomness of $f_s$ induces multiple possible sample paths starting from the same $\hat{\mathbb P}_0$. The bound~\eqref{eq:recipr-distortion} then holds \emph{almost surely pathwise in $\hat{\mathbb P}_T$}; it applies to every path along which $f_s$ is $L_s$-Lipschitz---which, by Condition~\ref{cond:lipschitz-sample-adpat}, are all paths, outside a null set. The possible values for $\hat{\mathbb P}_T$ that violate the bound~\eqref{eq:recipr-distortion} thus have probability zero. %
In later results, we will make similar qualifiers---that bounds hold pointwise in $\hat{\mathbb P}_0$, or with some probability over $\hat{\mathbb P}_0$, and almost surely pathwise in $\hat{\mathbb P}_T$. The meaning of those statements is exactly the same as described here.

Lemma~\ref{lemma:recipr-distortion} tells us how far in Wasserstein space the initial sample can be distorted by reciprocal learning. Thus, it will serve as a pivotal instrument in deriving generalization bounds later. Apart from the diameter $\diamZ$ of the instance space, which is a fixed quantity, the reciprocal distortion bound depends on the Lipschitz constant $L_s$ of the sample adaptation function, the initial sample size $n$, the number $m$ of points changed by the first sample adaptation and the iteration $T$ (i.e., the number of sample adaptations). 

For a given iteration $T \geq 2$, this bound grows in $L_s$. %
In other words, the less continuous the sample adaptation is, the further away it can potentially shift the initial empirical distribution in Wasserstein space. %
Simple calculations show that the bound's growth is superlinear in $T$ for $L_s > 1$, linear for $L_s = 1$, and sublinear for $L_s < 1$. %
In the latter case, %
the bound goes to 
\begin{equation*}%
\beta_\infty = \frac{1}{1 - L_s} \left(\frac{m }{n}\right)^{\frac{1}{p}} \diamZ ,    
\end{equation*}
as $T \rightarrow \infty$. We can see that this expression still grows with $L_s$. %
The effect of the initial sample size $n$ on the bound in Lemma~\ref{lemma:recipr-distortion} has an equally intuitive interpretation. %
It can be understood as the degree of idleness (or sluggishness) in reciprocal learning's first iteration. %
The bigger the initial sample is (fixing $m$), the more \say{sluggish} the maximal change in this sample 
is. %
That is to say, the larger the sample is, the harder it is for a reciprocal learning algorithm to move that sample in Wasserstein space. 

The sufficient conditions in \citet{rodemann24reciprocal} for $f_s$ to be Lipschitz imply that $L_s = (n-1)/n$ %
in the simple case of binary classification with a greedy sample adaptation function that adds exactly one data point per iteration and removes none \citep[see Theorem 1 of][]{rodemann24reciprocal}. In this case, $\beta_T \rightarrow 0$ as $n \rightarrow \infty$ and $\beta_\infty = m \mathscr{D}_{\mathcal{Z}}$ for $p=1$ and any $n$. %

Lemma~\ref{lemma:recipr-distortion} gives rise to a reciprocal ambiguity set
\begin{equation*}%
    \mathcal{A}_{\beta_T}(\hat{\mathbb P}_0) = 
    \left\{Q \in \mathcal{P}(\mathcal Z) : W_{p}(\hat{\mathbb P}_0, Q) \leq  \beta_T %
\right\}.
\end{equation*}
This ambiguity set is not chosen \textit{a priori} to robustify learning as in distributionally robust optimization. Instead, it characterizes the behavior of reciprocal learning algorithms \textit{a posteriori} because it is a function of the realized initial sample. %
The adapted sample $\hat{\mathbb P}_T$ must be in $\mathcal{A}_{\beta_T}(\hat{\mathbb P}_0)$ by Lemma~\ref{lemma:recipr-distortion}. %
In order to make statements on how well reciprocal learning generalizes beyond training data, we also need to consider the Wasserstein distance between $\hat{\mathbb P}_{T}$ and the true law $\mathbb P$. Our strategy in this regard will be to exploit the following corollary of \citet{fournier2015rate}, which bounds the Wasserstein distance between the empirical distribution $\hat{\mathbb P}_0$ and its population law $\mathbb P$ (as seen in Figure~\ref{fig:illu-wasserstein-space}). We use this bound together with Lemma~\ref{lemma:recipr-distortion} to bound the distance between $\hat{\mathbb P}_{T}$ and $\mathbb P$ (as illustrated in Figure~\ref{fig:illu-bounds}). %

\begin{lemma}[Convergence in Wasserstein Space]\label{lemma-wasserstein-iid}\label{lemma:wasserstein-conv}
Let $\hat{\mathbb P}_0$ denote the empirical distribution of $Z_{1}, \ldots, Z_{n} \iidsim \mathbb P$. Assume $ \dz > 2p$. Then, for any $\beta_0 \in(0, \infty)$,
\begin{equation}\label{eq:wasserstein-conv}
    W_{p}\left(\hat{\mathbb P}_0, \mathbb P\right) \leq \beta_0,
\end{equation}
with probability at least $1- C_{a} \exp \left(-C_{b} n \beta_0^{\dz / p}\right) $,
where $C_{a}$ and $C_{b}$ are positive constants %
which depend on $p, \dz$ and $\diamZ$ only.
\end{lemma}
Lemma~\ref{lemma:wasserstein-conv} is an application of \citeauthor{fournier2015rate}'s Theorem 2 (Case 1) with $\mathcal{E}_{\alpha, \gamma}(\mu) < \infty $ and  $\alpha = \dz > p$. This mirrors how \citeauthor{fournier2015rate}'s Theorem 2 is used in Proposition 5 of \citet{lee2018minimax}. %
It implies that $\hat{\mathbb P}_0$ is arbitrarily close to $\mathbb P$ in Wasserstein space with high probability as $n \rightarrow \infty$. %
Note that \citet{fournier2015rate} also show that the bound \eqref{eq:wasserstein-conv} still holds for $\dz \leq 2p$, but with slower rate of approximation, which would render our bounds wider. We will assume $\dz > 2p$ throughout the remainder of the paper. %

\subsection{Generalization Bounds for Convergent Reciprocal Learning}\label{sec:bounds-convergent-sol}

We are now ready to give generalization bounds that answer the questions of whether and how well reciprocal learning can generalize from its adaptively selected sample.
We start with our main results for convergent reciprocal learning algorithms, before moving to the finite-iteration case in the next section. These results bound the generalization error $\mathscr R(\mathbb P, \hat{\theta}_c)$ of reciprocal learning's output $\hat \theta_c$. 
The above-mentioned sufficient criteria for convergence in \cite{rodemann24reciprocal}
guarantee that $\hat \theta_c$ is (pathwise) unique for a given reciprocal learning algorithm and given sample $\hat{\mathbb{P}}_0$. We can thus identify any reciprocal learning algorithm fulfilling these criteria with its convergent solution $\hat \theta_c$ and bound its generalization error via the following theorem. Moreover, the theorem also holds under any other sufficient conditions for convergence.

\begin{theorem}[Generalization Gap]\label{thm:gen-error-bound}    
        Suppose the loss is $L_{\ell}$-Lipschitz in the data (Condition~\ref{ass-cont-diff-feat-param-Lips-data}) and reciprocal learning converges to $(\hat{\theta}_c, \hat{\mathbb P}_c)$ (Condition~\ref{cond:conv-reciprocal-learning}) with $L_s$-Lipschitz sample adaptation (Condition~\ref{cond:lipschitz-sample-adpat}) where $L_s < 1$. Then

    \begin{equation}\label{eq:thm:gen-error-bound}
            \mathscr R(\mathbb P, \hat{\theta}_c) \leq  \mathscr R(\hat{\mathbb P}_c, \hat{\theta}_c) + L_\ell \left(\frac{\log \left(C_a / \delta\right)}{C_{b} n}\right)^{p / d_z} + \frac{ L_\ell \left(\frac{m }{n}\right)^{\frac{1}{p}} \diamZ}{1 - L_s} ,

    \end{equation}
        with probability at least $1-\delta$%
    , where $C_a$ and $C_b$ are constants (from Lemma~\ref{lemma-wasserstein-iid}) which depend on $p$, $\dz$ and $\diamZ$. %

\end{theorem}
Recall that $\diamZ$ is the diameter of $\mathcal Z$ which is finite by the assumption that $\mathcal Z$ is bounded. The main idea of Theorem~\ref{thm:gen-error-bound}'s proof, as detailed in Appendix~\ref{app:proofs}, is illustrated by Figures~\ref{fig:illu-wasserstein-space} and~\ref{fig:illu-bounds}. We combine the bound from Lemma~\ref{lemma:wasserstein-conv} on the $p$-Wasserstein distance between the law $\mathbb P$ and an initial \iid\ sample $\hat{\mathbb P}_0$ with the bound from Lemma~\ref{lemma:recipr-distortion} on how far reciprocal learning algorithms move that initial \iid\ sample $\hat{\mathbb P}_0$ in Wasserstein space to $\hat{\mathbb P}_c$. %
Equation~\ref{eq:thm:gen-error-bound} holds almost surely pathwise, while the probability $1-\delta$ in Theorem~\ref{thm:gen-error-bound} arises from the \iid\ sampling of $\hat{\mathbb P}_0$ from $\mathbb P$. Such is the case with all our results that rely on Conditions~\ref{cond:conv-reciprocal-learning} or~\ref{cond:lipschitz-sample-adpat}---they hold almost surely over the reciprocal paths. In contrast, all the probability guarantees (such as $1-\delta$ or $1-2\delta$) in our results are with respect to the randomness in $\hat{\mathbb P}_0$.

Theorem~\ref{thm:gen-error-bound}, as with all our results, holds for any choice of $p$ which satisfies its assumptions. (Apart from the requirement $1 \le p < 0.5 d_z$ which is assumed throughout the paper, $p$ also appears in Condition~\ref{cond:lipschitz-sample-adpat}.) Hence $p$ can be optimized to find the tightest bounds implied by our results. However, optimizing $p$ often requires a nontrivial tradeoff. For example, in Theorem~\ref{thm:gen-error-bound}, the generalization gap's third term grows in $p$ while the second term decreases in $p$ (assuming that $\left(\frac{\log \left(C_a / \delta\right)}{C_{b} n}\right)^{1 / d_z} < 1$). This means that the optimal $p$ must be determined on a case-by-case basis.

Theorem~\ref{thm:gen-error-bound} gives us a guarantee on how much worse---in comparison to its training error---our reciprocal learner $\hat{\theta}_c$ could perform on unseen test data. When every predictor $f_\theta$ performs poorly (in the sense that its risk $\mathscr R(\mathbb P, \theta)$ is large), this is an unfair evaluation of reciprocal learning because no matter how reciprocal learning selects $\hat \theta_c$, it will have a large generalization error. 
A more meaningful and general comparison is between the risk of $\hat\theta_c$ and the infimum risk over the hypothesis class $\mathcal F$ (i.e., the Bayes risk).
For this comparison, we have to consider the excess risk 
$ \mathscr R({\mathbb P}, \hat{\theta}_c) - \mathscr R_\Theta $.  
While a small generalization error tells us that our learner’s performance on the training set generalizes well, a small excess risk indicates that the risk of our learner's predictor is close to the best achievable risk.

\begin{theorem}[Data-Dependent Excess Risk Bound]\label{thm:gen-error-bound-data-dependent}  
        Suppose the loss is $L_{\ell}$-Lipschitz in the data (Condition~\ref{ass-cont-diff-feat-param-Lips-data}) and reciprocal learning converges to $(\hat{\theta}_c, \hat{\mathbb P}_c)$ (Condition~\ref{cond:conv-reciprocal-learning}). Then the excess risk $\mathscr R({\mathbb P}, \hat{\theta}_c) - \mathscr R_\Theta $ is upper bounded by

    \begin{equation*}%
                \mathscr R(\hat{\mathbb P}_0, \hat{\theta}_c) - \mathscr R(\hat{\mathbb P}_0, \hat{\theta}_0) + L_\ell \left(\frac{\log \left(C_a / \delta\right)}{C_{b} n}\right)^{p / d_z}  +
        \frac{24}{\sqrt n} \mathfrak C_{L_2}(\mathcal L_0) + L_\ell \diamZ \sqrt{\frac{\log(1/\delta)}{2n}},

    \end{equation*}
        with probability at least $1-2\delta$.

\end{theorem}
Here, as elsewhere, $\mathcal L_0 $ %
is the class of centered loss functions %
(Equation~\ref{eq:defnCenteredLossFunctionClass}) and $\mathfrak C_{L_2}(\mathcal L_0)$ is its covering entropy integral (see Definition~\ref{def:covering-n}). When we can write the loss $\ell$ in terms of $f_\theta$ rather than $\theta$, the covering entropy integral $\mathfrak C_{L_2}(\mathcal L_0)$ can be replaced by a complexity measure involving the hypothesis class $\mathcal F$, as we show in the following corollary.

\begin{cor}\label{cor:gen-error-bound-data-dependent}
        Assume that the loss function $\ell$ takes the form
    \[\ell(\theta, x, y) = \tilde \ell \bigl( f_\theta(x), y \bigr) + r(\theta),\]
    where $r : \Theta \to \mathbb R$ is an arbitrary regularizer and $\tilde \ell : \mathcal Y \times \mathcal Y \to \mathbb R$ is the unpenalized loss function. (We place no restrictions on the form of $r$; in particular, an unpenalized loss $\ell$ can be handled by taking $r(\theta) = 0$.) Further suppose that $\tilde \ell (\,\cdot\,, y')$ is $L_\ell$-Lipschitz for all $y' \in \mathcal Y$. Then we may replace $\mathfrak C_{L_2}(\mathcal L_0)$ with $L_{\ell} \mathfrak C_{L_2}(\mathcal F)$ in each of the excess risk bounds---that is, in Theorems~\ref{thm:gen-error-bound-data-dependent},~\ref{thm:excess-risk-classic},~\ref{thm:excess-risk-classic-anytime} and~\ref{thm:data-dependent-anytime-excess}.

\end{cor}
Unlike in Theorem~\ref{thm:gen-error-bound}, we do not require the sample adaptation to be Lipschitz continuous in Theorem~\ref{thm:gen-error-bound-data-dependent}, nor in the corresponding result given by Corollary~\ref{cor:gen-error-bound-data-dependent}. %
However, the bounds in Theorems~\ref{thm:gen-error-bound} and~\ref{thm:gen-error-bound-data-dependent} are data dependent (via $\mathscr R(\hat{\mathbb P}_c, \hat{\theta}_c)$ in the former and via $\mathscr R(\hat{\mathbb P}_0, \hat{\theta}_c) - \mathscr R(\hat{\mathbb P}_0, \hat{\theta}_0)$ in the latter). %
In other words, they use information from the realized sample. It is natural therefore to ask whether we can also provide universal data-independent bounds along the lines of classical empirical process theory \citep{vaart1997weak}. The following theorem gives an affirmative answer. %

\begin{theorem}[Excess Risk Bound]\label{thm:excess-risk-classic}
       Suppose $\Theta$ is closed and convex. Assume the loss is continuously differentiable and $L_{\ell}$-Lipschitz (Condition~\ref{ass-cont-diff-feat-param}) as well as strongly convex (Condition~\ref{ass-str-convex}). If reciprocal learning converges to $(\hat{\theta}_c, \hat{\mathbb P}_c)$ (Condition~\ref{cond:conv-reciprocal-learning}) with $L_s$-Lipschitz sample adaptation (Condition~\ref{cond:lipschitz-sample-adpat}), where $L_s < 1$, then its excess risk
    $ \mathscr R({\mathbb P}, \hat{\theta}_c) - \mathscr R_\Theta $ is upper bounded by
    \[
        L_\ell \left(\frac{\log \left(C_a / \delta\right)}{C_{b} n}\right)^{p / d_z} +  \frac{L_\ell  (\kappa/\gamma) \left(\frac{m }{n}\right)^{\frac{1}{p}} \diamZ}{ 1 - L_s } +
        \frac{24}{\sqrt n} \mathfrak C_{L_2}(\mathcal L_0) + L_\ell \diamZ \sqrt{\frac{\log(1/\delta)}{2n}},
    \]
    with probability at least $1-2\delta$.

\end{theorem}
Recall that $\kappa$ and $\gamma$ are constants given in Conditions~\ref{ass-cont-diff-feat-param-cts-diff-param} and~\ref{ass-str-convex} respectively, and %
$L_\ell$ is the Lipschitz constant of the loss. 
As in Theorems~\ref{thm:gen-error-bound} and~\ref{thm:gen-error-bound-data-dependent}, $C_a$ and $C_b$ are constants which depend only on $p, \dz$ and $\diamZ$. 

A complete proof of Theorem~\ref{thm:excess-risk-classic} is in Appendix~\ref{proof:excess-risk-classic}. Its main idea is as follows: %
We first link the generalization error of $\hat \theta_c$ to the risk on the initial \iid\ sample $\hat{\mathbb P}_0$. Secondly, we compare this risk to that of the initial one-shot ERM $\hat{\theta}_0$. Finally, leveraging the fact that the difference between this risk and the Bayes risk $\mathscr R_\Theta$ is a classical empirical process allows us to exploit empirical process theory to derive bounds on this excess risk.

By means of Theorem~\ref{thm:excess-risk-classic}, we can bound the excess risk of any convergent reciprocal learning algorithm. The bound is a function only of the centered loss class $\mathcal L_0$ (or of the hypothesis class $\mathcal F$ in the corresponding result given by Corollary~\ref{cor:gen-error-bound-data-dependent}) and some other problem-dependent constants, but does not depend on the realized sample. 

Both the generalization gap (Theorem~\ref{thm:gen-error-bound}) and the excess risk bounds (Theorems~\ref{thm:gen-error-bound-data-dependent} and~\ref{thm:excess-risk-classic}) have intuitive interpretations. As a first observation, the bounds in Theorems~\ref{thm:gen-error-bound} and~\ref{thm:excess-risk-classic} grow with $L_s$. The bigger $L_s$ is, the more sensitive reciprocal learning is towards previous changes in the sample and thus the farther away $\hat{\mathbb P}_{t}$ can potentially move away from $\hat{\mathbb P}_0$ in the worst case. When $L_s$ approaches zero, the sample adaptation function approaches the \say{smoothest} function---namely the constant function. More generally, smaller $L_s$ implies less potential variation in the sample adaptation, thus distorting the  initial sample $\hat{\mathbb P}_0$ less. This is because $L_s$ restricts the change in the sample relative to the previous change in the sample; that is to say, $L_s$ is the maximum amount of change (from $\hat{\mathbb P}_t$ to $\hat{\mathbb P}_{t+1}$) that the previous change (from $\hat{\mathbb P}_{t-1}$ to $\hat{\mathbb P}_t$) can induce. (The smaller $L_s$ is, the less sensitive $f_s$ is to previous changes in the sample.) %
As mentioned in the discussion after Lemma~\ref{lemma:recipr-distortion}, we have $L_s = (n-1)/n$ for binary classification under some conditions on the loss and the sample adaptation function (specifically, only $m_t=1$ data points being added per iteration). %
When $p > 1$, this implies that the third term of Theorem~\ref{thm:gen-error-bound}'s bound grows in $n$. In contrast, when $p = 1$, $L_s = (n-1)/n$ implies that this bound only depends on $n$ through its second term, which is derived from the classical bound on the Wasserstein distance between the true law $\mathbb P$ and the \iid\ sample $\hat{\mathbb P}_0$ (Lemma~\ref{lemma:wasserstein-conv}). %
In this scenario, reciprocal learning (in its worst case) merely adds the term $L_\ell \mathscr D_{\mathcal Z}$ %
to ERM's classical generalization gap. %

Moreover, the bounds in Theorems~\ref{thm:gen-error-bound} and \ref{thm:excess-risk-classic} decrease as the sample size $n$ increases, and vanish as $n \to \infty$, provided that $L_s$ does not depend on $n$ and $m = o(n)$. (Theorem~\ref{thm:gen-error-bound-data-dependent}'s bound does not necessarily decrease with $n$, nor vanish as $n \to \infty$ because it contains the data-dependent term $\mathscr R(\hat{\mathbb P}_0, \hat{\theta}_c) - \mathscr R(\hat{\mathbb P}_0, \hat{\theta}_0)$, whose dependence on $n$ we do not characterize. Apart from this term, the rest of Theorem~\ref{thm:gen-error-bound-data-dependent}'s bound is decreasing in $n$ and vanishes as $n \to \infty$.) This is equally intuitive. While $L_s$ accounts for the movement in Wasserstein space \textit{relative} to the previous movement, the sample size $n$ determines the \textit{absolute} inertia of the bound's growth. %
As discussed for Lemma~\ref{lemma:recipr-distortion}, $n$ can be understood as the degree of idleness in reciprocal learning's first iteration: The higher $n$, the harder it is for the reciprocal learning algorithm to move the initial sample in Wasserstein space---that is, the more sluggish the bound's growth in the iterations becomes%
. As our generalization bounds account for the worst case, this idleness makes them tighter. 

We observe that all bounds grow in the Lipschitz constant $L_\ell$ of the loss function. Intuitively, a higher $L_\ell$ means that a given movement (of the sample in Wasserstein space, or of the fitted parameter) can result in a larger change in the risk.
The excess risk bound in Theorem~\ref{thm:excess-risk-classic} additionally depends on $\kappa/\gamma$, which is a Lipschitz constant for the ERM map $ P \mapsto \argmin_{\theta \in \Theta} \mathscr R(P, \theta)$ from $(\mathcal{P}(\mathcal Z), W_p)$ to $(\Theta, \| \cdot \|_2)$ under Conditions~\ref{ass-cont-diff-feat-param} and~\ref{ass-str-convex} (see Appendix~\ref{proof:excess-risk-classic} for a proof of this fact). %
The larger $\kappa/\gamma$ is, the further the fitted parameter can move for a given distortion of the sample.

\subsection{Anytime-Valid Generalization Bounds for Reciprocal Learning}\label{sec:bounds-anytime}

The previous section analyzed the generalization performance of the limiting parameter $\hat \theta_c$ of a convergent reciprocal learning algorithm. It is natural to ask whether one can obtain generalization bounds that hold for the fitted parameter $\hat \theta_T$ rather than for the convergent solution $\hat \theta_c$. This question is of high practical relevance. By Theorem~\ref{thm:convergence}, %
practitioners can check whether their reciprocal learning algorithm converges. Section~\ref{sec:bounds-convergent-sol} then allows them to %
assess the generalization guarantees of $\hat{\theta}_c$. While this provides a principled analysis of any reciprocal learning algorithm that is run until (approximate) convergence, practitioners might also be interested in other algorithms---either those that do not converge, or those that are stopped early at some iteration $T$. This is because a machine learning operator might simply apply a reciprocal learning algorithm---convergent or not---and stop at some iteration~$T$. Whether the algorithm would have converged or not if it had continued to run is of little practical relevance in this scenario; equally irrelevant is the generalization performance of $\hat \theta_c$ (assuming it even exists). In this situation, all the operator is interested in is: \textit{How well does my reciprocal learning algorithm generalize if I were to stop it now at iteration~$T$?}   

This section answers this question via the following Theorem~\ref{thm:gen-error-bound-anytime} and subsequent Theorems~\ref{thm:excess-risk-classic-anytime} and~\ref{thm:data-dependent-anytime-excess}. Notably, all three theorems require only a subset of the conditions used in the previous results. In particular, Theorem~\ref{thm:gen-error-bound-anytime} does not require the loss function to be strongly convex, making it applicable to a wide range of loss functions, including those of neural networks. Moreover, Condition~\ref{cond:conv-reciprocal-learning} (convergence) is no longer needed either, opening up new applications. Convergence has only been proven for reciprocal learning algorithms with non-greedy sample adaptation (see Theorem~\ref{thm:convergence}), while the results in this section can be applied to both greedy and non-greedy sample adaptation functions.

\begin{theorem}[Anytime-Valid Generalization Gap]\label{thm:gen-error-bound-anytime}   
        Let $\hat \theta_{T}$ be the ERM solution at iteration $T$. If Conditions~\ref{ass-cont-diff-feat-param-Lips-data} and~\ref{cond:lipschitz-sample-adpat} hold, then

    \begin{equation}\label{eq:gen-error-bound-anytime}
            \mathscr R(\mathbb P, \hat{\theta}_T) \leq   \mathscr R(\hat{\mathbb P}_T, \hat{\theta}_T) + L_\ell \left(\frac{\log \left(C_a / \delta\right)}{C_{b} n}\right)^{p / d_z} +    L_\ell S_T(L_s) \left(\frac{m }{n}\right)^{\frac{1}{p}} \diamZ,

    \end{equation}
        for any $T$, with probability at least $1-\delta$, where $S_T(L_s)$ is the geometric sum defined in Equation~\ref{eq:defn-geo-sum}.%

\end{theorem} %
Importantly, the above generalization gap holds for all iterations $T$ simultaneously with probability $1-\delta$ (in comparison to the gap holding with this probability for each $T$ individually). Since this means the generalization gap remains valid at all stopping times, this justifies our use of the term ``anytime valid'' \citep{ramdasGameTheoreticStatisticsSafe2023} 
and allows for the construction of stopping rules along the lines of the following corollary:

\begin{cor}[Generalization Under Stopping Rules]\label{cor_stop}
        Suppose that Conditions~\ref{ass-cont-diff-feat-param-Lips-data} and \ref{cond:lipschitz-sample-adpat} hold. Let $T$ be a stopping rule (or indeed a possibly data-dependent, random iteration). If we stop reciprocal learning at iteration $T$, the generalization gap bound (Equation~\ref{eq:gen-error-bound-anytime}) holds
    with probability at least $1 - \delta$.

\end{cor}
Analogous stopping rule corollaries also hold for the other results in this section (Theorems~\ref{thm:excess-risk-classic-anytime} and \ref{thm:data-dependent-anytime-excess}) by the same argument as above.

Just as in Section~\ref{sec:bounds-convergent-sol}, it is natural to ask whether one can derive universal data-independent bounds on the excess risk at any iteration $T$. The following theorem answers this question in the affirmative by bounding %
$\mathscr R({\mathbb P}, \hat{\theta}_T) - \mathscr R_\Theta $ under the same conditions as in Theorem~\ref{thm:gen-error-bound-anytime} with the additional requirements of Condition~\ref{ass-cont-diff-feat-param-cts-diff-param} and a strongly convex loss.

\begin{theorem}[Anytime-Valid Excess Risk Bound]\label{thm:excess-risk-classic-anytime}
        Assume that $\Theta$ is closed and convex. %
    Given Conditions~\ref{ass-cont-diff-feat-param},~\ref{ass-str-convex} and~\ref{cond:lipschitz-sample-adpat}, the excess risk
    $ \mathscr R({\mathbb P}, \hat{\theta}_T) - \mathscr R_\Theta $ is upper bounded by
    \[
         L_\ell \left(\frac{\log \left(C_a / \delta\right)}{C_{b} n}\right)^{p / d_z}
        + L_\ell  S_T(L_s) \frac{\kappa}{\gamma} \left(\frac{m }{n}\right)^{\frac{1}{p}} \diamZ
        + \frac{24}{\sqrt n} \mathfrak C_{L_2}(\mathcal L_0) + L_\ell \diamZ \sqrt{\frac{\log(1/\delta)}{2n}},
    \]
    for any stopping rule $T$, with probability at least $1-2\delta$.%

\end{theorem}
Further, we can derive a data-dependent anytime-valid excess risk bound by adapting the proof of the previous theorem.

\begin{theorem}[Data-Dependent Anytime-Valid Excess Risk Bound]\label{thm:data-dependent-anytime-excess}
        Given Condition~\ref{ass-cont-diff-feat-param-Lips-data}, the excess risk
    $ \mathscr R({\mathbb P}, \hat{\theta}_T) - \mathscr R_\Theta $ is upper bounded by
    \[
        \mathscr R(\hat{\mathbb P}_0, \hat{\theta}_T) - \mathscr R(\hat{\mathbb P}_0, \hat{\theta}_0) + L_\ell \left(\frac{\log \left(C_a / \delta\right)}{C_{b} n}\right)^{p / d_z}
        +
        \frac{24}{\sqrt n} \mathfrak C_{L_2}(\mathcal L_0) + L_\ell \diamZ \sqrt{\frac{\log(1/\delta)}{2n}},
    \]
    for any stopping rule $T$, with probability at least $1-2\delta$.%

\end{theorem}
This bound only requires Condition~\ref{ass-cont-diff-feat-param-Lips-data}. 
The price we pay for this relatively weak requirement is a practical one: Using the bound as a stopping criterion in reciprocal learning requires computing $\mathscr R(\hat{\mathbb P}_0, \hat{\theta}_T)$ in each iteration $T$. However, note that this translates to an additional model evaluation of $\hat \theta_T$ on $\hat{\mathbb P}_0$, and not a model refit.  

An important distinction between Theorem~\ref{thm:data-dependent-anytime-excess} and the data-independent bounds in Theorems~\ref{thm:gen-error-bound-anytime} and~\ref{thm:excess-risk-classic-anytime} is their dependence on the iteration $T$. The latter bounds contain the worst-case reciprocal-distortion term $S_T(L_s)$, which is nondecreasing in $T$ (and strictly increasing when $L_s>0$). By contrast, the $T$-dependent term in Theorem~\ref{thm:data-dependent-anytime-excess} is the observed quantity $\mathscr R(\hat{\mathbb P}_0, \hat \theta_T)- \mathscr R(\hat{\mathbb P}_0,\hat \theta_0)$. Although this quantity is nonnegative (because $\hat\theta_0$ is an empirical risk minimizer on $\hat P_0$), it need not increase monotonically with $T$. Consequently, the data-dependent bound can remain constant or even decrease across iterations, thereby adapting to reciprocal-learning paths that distort performance on the initial sample less than the worst-case analysis permits. 
In this way, our analysis captures cases where reciprocal learning corrects itself---more specifically, cases where a method improves upon the risk of earlier $\hat{\theta}_t$ (with respect to $\hat{\mathbb P}_0$).

\section{Example: Stopping Rules for Semi-Supervised Learning}\label{sec:example}

\begin{figure}
    \centering
    \includegraphics[width=0.82\linewidth]{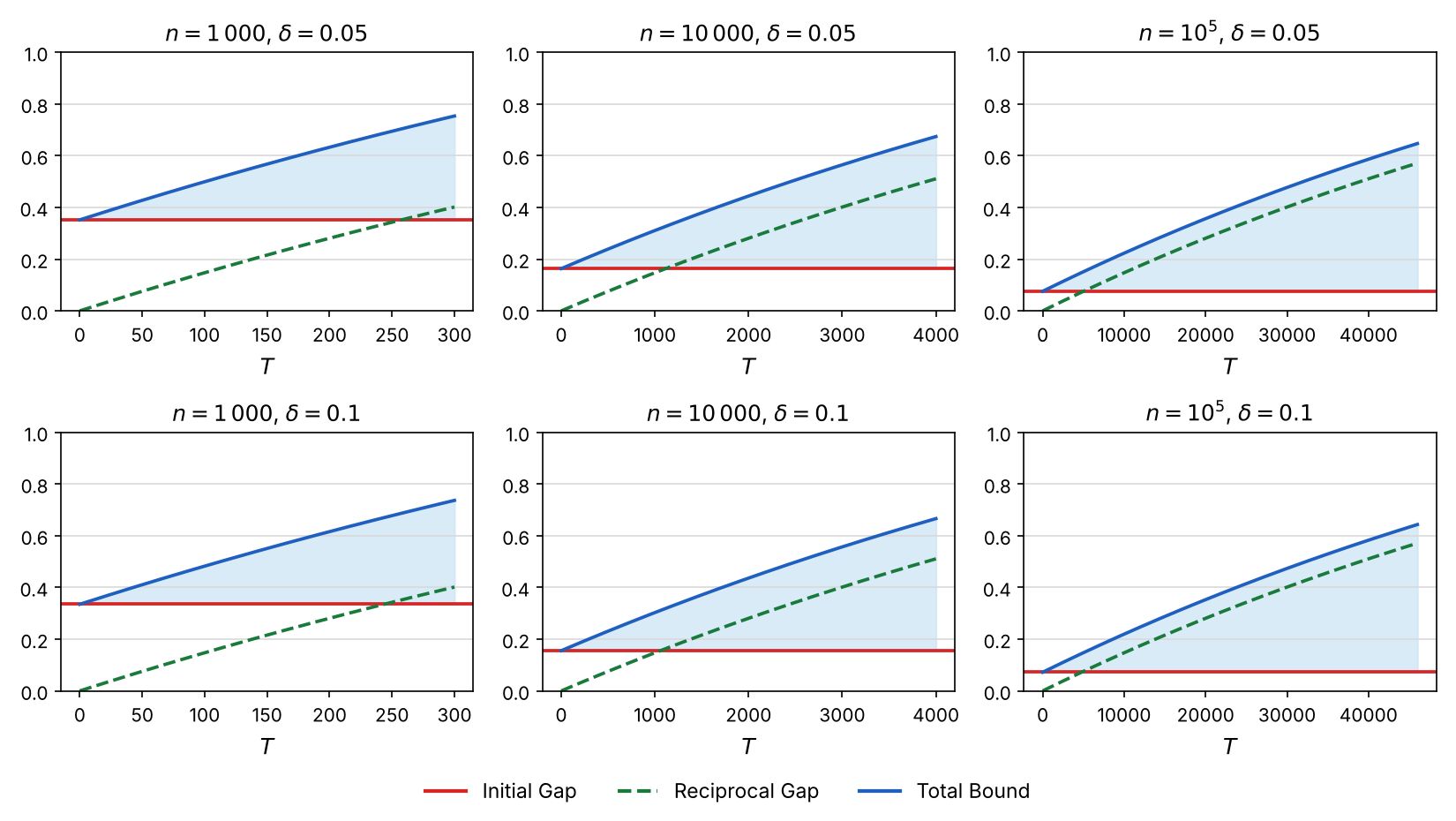}
    \caption{%
    An illustration of the generalization gap bound from Equation~\ref{eq:bound-example} with $\delta \in \{0.05,0.1\}$, i.e., the bounds hold with probability at least $0.95$ and $0.9$, respectively. The plots show the total bound (solid blue) as well as its two components: the initial gap (solid red) and the reciprocal gap (dotted green).}
    \label{fig:simple-bounds}
\end{figure}

In order to get a sense of what our generalization bounds look like in practice, let us consider a very simple toy example for the sake of illustration. %
Consider semi-supervised learning via self-training for binary  classification ($\mathcal{Y} = \{0,1\}$) with $\mathcal{X} \subset \mathbb R^2$ and $\Theta \subset \mathbb R^2$, as depicted in Figure~\ref{fig:self-training}. (This implies $\mathcal{Z}$ has dimension $\dz= \dx + \dy = 3$.) For instance, suppose we are predicting pipeline failure ($y = 1$) from standardized continuous feature vectors $x \in \mathcal{X} = [0,1]^2$, say \texttt{pipe age} and \texttt{operational pressure} as in \citet[Table 1]{alobaidi2022semi}. If we want to ensure the generalization error of our failure predictions is not worse than, e.g., $0.2$ above  the training error we measure, we can use our stopping rules to guarantee that.
Assume we are using the binary logistic loss $
\ell(\theta, x,y) =\log\bigl(1+\exp(\langle\theta,x\rangle)\bigr)
-y\langle\theta,x\rangle
$ and let $\Theta=[-2/5,2/5]^2$. Writing
$s=\langle\theta,x\rangle$ and
$\sigma(s)=(1+e^{-s})^{-1}$, we have
$ \nabla_{(x,y)}\ell=\bigl((\sigma(s)-y)\theta,\,-s\bigr).
$ For $x\in[0,1]^2$ and $\theta\in\Theta$, the inner product $s$ is in $[-4/5,4/5]$, so that, for
$y\in[0,1]$,
$ \|\nabla_{(x,y)}\ell\|_2^2
\leq
\sigma(4/5)^2\frac{8}{25}+\frac{16}{25}<\frac{4}{5}.
$ Hence Condition~\ref{ass-cont-diff-feat-param-Lips-data} holds with
$L_\ell\le 2/\sqrt{5}$. (Note that Theorem~\ref{thm:gen-error-bound-anytime} requires Lipschitz continuity of the loss in the data only, so that the gradient with respect to $\theta$ plays no role here.) %
Further assume that the sample is modified according to a regularized Lipschitz-continuous uncertainty measure as in \citet{rizve2020defense} or \citet{arazo2020pseudo}, such that the sufficient conditions from \citet{rodemann24reciprocal} for the sample adaptation $f_s$ to be $L_s$-Lipschitz are fulfilled. Moreover, assume only $m=1$ data point is changed per iteration such that $L_s = (n-1)/n$. %
Thus, the assumptions of Theorem~\ref{thm:gen-error-bound-anytime} hold: Condition~\ref{ass-cont-diff-feat-param-Lips-data} was verified above; Condition~\ref{cond:lipschitz-sample-adpat} is assumed; and $\dz=3>2=2p$. Note that strong convexity and convergence are not required. Simple calculus shows our anytime-valid bound on the generalization gap from Theorem~\ref{thm:gen-error-bound-anytime} becomes %
\begin{equation}\label{eq:bound-example}
\mathscr R(\mathbb P, \hat{\theta}_T) -   \mathscr R(\hat{\mathbb P}_T, \hat{\theta}_T) \leq \underbrace{\frac{2}{\sqrt{5}} \left(\frac{12 \log \left(8 / \delta\right)}{n}\right)^{1 / 3}}_{\text{initial gap}} +  \underbrace{\frac{2\sqrt{15}}{5} \left(1- \left( \frac{n-1}{n}\right)^T\right)}_{\text{reciprocal gap}}
\end{equation}
with probability at least $1 - \delta$, where we used the Wasserstein-1 distance and %
constants $C_a = 2^{\dz/p} = 8$ and $C_b = 1 /(4 \mathscr D_\mathcal{Z}^2) = 1/12$, as well as $\mathscr D_\mathcal{Z} = \sqrt{3}$ (since $\mathcal Z = [0,1]^2 \times \{0,1\}$) and $L_\ell \le 2/\sqrt{5}$ as verified above.
Figure~\ref{fig:simple-bounds} shows how this bound---which consists of a bound on the initial gap and a bound on the reciprocal gap---grows in $T$ for different sizes $n$ of initial training samples and $\delta \in \{0.05,0.1\}$. 
The generalization gap bound can be interpreted as follows. In the case of, for example, $n = \num{10000}$, the generalization error of the model $\hat{\theta}_{100}$ (i.e., after adding $100$ pseudo-labeled data points) is at most
\[
\underbrace{\frac{2}{\sqrt{5}} \left(\frac{12 \log \left(8 / 0.05\right)}{\num{10000}}\right)^{1 / 3}}_{\text{initial gap}} +  \underbrace{\frac{2\sqrt{15}}{5} \left(1- \left( \frac{9999}{\num{10000}}\right)^{100} \right)}_{\text{reciprocal gap}} = \underbrace{0.1633}_{\text{initial gap}} +  \underbrace{0.0154}_{\text{reciprocal gap}} = 0.1788
\]
higher than its training error with $95\%$ confidence. Note that the initial gap can potentially be made smaller by different choices of $C_a$ and $C_b$ \citep[see][]{fournier2015rate}, as is common for such concentration-of-measure inequalities.

Since the training error is available to practitioners, this bound directly informs them about when to stop a reciprocal learning algorithm in order not to exceed a specific generalization error. Such a stopping rule can easily be derived by solving Equation~\ref{eq:bound-example} for $T$ after plugging in a desired maximal generalization error and the observed training error. In our toy example, we obtain the stopping rule: 
\begin{equation*}
T \leq \log\left( 1 - \frac{ 0.2  - \frac{2}{\sqrt{5}} \left(\frac{12 \log \left(8 / \delta\right)}{n}\right)^{1 / 3} }{2\sqrt{15}/5} \right) \big / \log\left( \frac{n-1}{n} \right)  = 239.48.
\end{equation*}
Thus, if we want to ensure the generalization gap $\mathscr R(\mathbb P, \hat{\theta}_T) - \mathscr R(\hat{\mathbb P}_T, \hat{\theta}_T)$ is at most $0.2$ with probability $0.95$, we must stop the algorithm before iteration $240$. %
Or, in other words, to guarantee that the generalization error of our failure predictions is not higher than $0.2$ above the training error, we ought to self-label no more than $239$ pipelines. %

These types of stopping rules with probabilistic guarantees are most useful in safety-critical applications. In addition to the example we have provided on oil and gas pipeline failure detection, our results could also prove useful in medical diagnosis, autonomous driving, aircraft engine monitoring, drug safety screening or cybersecurity intrusion detection, which all use some form of reciprocal learning. 

\section{Discussion and Outlook}\label{sec:conclusion}

Standard statistical learning theory provides generalization bounds for models trained by %
ERM using a given, fixed sample. In practice, however, ERM is often applied sequentially to changing samples. In the %
common %
case of reciprocal learning, the sequential model updates are dependent on each other, making generalization bounds nontrivial to obtain. In this work, we embed %
reciprocal learning into statistical learning theory, thereby %
extending the classical fixed-sample framework to feedback-driven learning. This allows us to derive universal generalization guarantees on the wide range of paradigms %
covered by reciprocal learning, including active learning, bandits, boosting, semi-supervised learning, Bayesian optimization and superset learning. In addition to generalization bounds for convergent reciprocal learning, we also derive anytime-valid bounds, which have the added advantage of giving rise to stopping criteria. %
We demonstrate the applicability of these stopping criteria to semi-supervised learning for oil and gas pipeline failure detection, but they are equally applicable to a wide range of other safety-critical domains.

On a high level, our analysis reveals that sample-changing learning algorithms can in fact reliably generalize, if the initial sample provides enough information about the population and the way the algorithm changes this sample is restricted. Interestingly, these restrictions are linked to the sufficient conditions for reciprocal learning's convergence given in \citet{rodemann24reciprocal}. %
Along with regularity conditions on the loss function, our generalization bounds require either that the reciprocal learner converges or that its sample adaptation function is Lipschitz continuous---with the notable exception of Theorem~\ref{thm:data-dependent-anytime-excess}, which only requires Lipschitz continuity of the loss. %
Theorems~\ref{thm:excess-risk-classic} and~\ref{thm:excess-risk-classic-anytime} further need the loss to be strongly convex---a condition which is also required by \citet{rodemann24reciprocal} for convergence.

Instead of discussing these conditions along a subjective weak--strong spectrum, we would like to emphasize the \textit{subject} of our conditions: the loss and the sample adaptation functions of the specific reciprocal learning algorithm in question. The drawback of these conditions is that they exclude a considerable number of algorithms within each reciprocal learning paradigm. For instance, the requirement of strongly convex loss for Theorems~\ref{thm:excess-risk-classic} and~\ref{thm:excess-risk-classic-anytime} excludes any kind of multilayer neural networks.   
On the other hand, however, a key advantage of this approach is that all our conditions are verifiable. Practitioners can check whether their algorithm fulfills the sufficient conditions for our generalization bounds to hold. %
They are not forced to base their conclusions on unverifiable assumptions regarding, for example, the label annotation process (in active learning), the pseudo-label's quality (in semi-supervised learning), or the reward function (in bandits), to name only a few conditions commonly required in the field. 

In terms of conditions on the data, our analysis only relies on the initial sample, which we assume to be drawn \iid\ from the true law. No assumptions about how the newly added data is distributed are required. This is a very natural perspective for practitioners in various domains: While the initial gold standard data is often trusted \citep{angelopoulos2023prediction,angelopoulos2023ppi++}, caution is warranted about newly acquired data due to, e.g., erroneous human labeling (in active learning) or biased reward functions (in bandits). In semi-supervised learning and boosting, this mistrust is even more pronounced, owing to the risk of self-referential inference: Both methods change the sample via their own model's predictions. Given that all models are wrong \citep{box1976science}, trusting adapted samples can be unwarranted. Our results reveal that inference is still possible, if we force the changes in the sample to be bounded in Wasserstein distance.    
Summing up, the appeal of our analysis lies in its twofold universality---covering a wide range of algorithms, while requiring no assumptions on the distribution of acquired or altered data.
Accordingly, we do not claim paradigm-specific optimality for special cases of reciprocal learning. Rather, our results provide a unified, assumption-lean and
algorithmically verifiable baseline for inference from adaptively modified samples, especially when stronger paradigm-specific assumptions are unavailable, undesirable, or unjustifiable.

This article opens up several avenues for future work. Most notably, our generalization bounds' sufficient conditions pave the way for the design of novel reciprocal learning algorithms with probabilistic generalization guarantees. %
On the one hand, this %
could involve modifying existing algorithms. For instance, regularizing data selection (e.g., the acquisition function in active learning) in order to render the sample adaptation function Lipschitz would allow the application of anytime-valid excess risk bounds along the lines of Section~\ref{sec:bounds-anytime}. This in turn opens the door to %
stopping criteria similar to the ones illustrated in Section~\ref{sec:example}, allowing for more reliable and trustworthy active learning---something %
that is crucial for any safety-critical application, like, for instance, identifying patient-safety events \citep[see][for an application of active learning to this problem]{fong2017using}. On the other hand, our generalization bounds' sufficient conditions also allow for a number of opportunities to develop novel, theory-informed reciprocal learning algorithms beyond the common examples of active learning, boosting, bandits, and the like. 

Such ideas do not even need to be restricted to machine learning algorithms. Longitudinal medical studies, for example, can be subject to feedback loops between the learned effects and the sample in the case of double-blind protocol violations
\citep{BANG2004143}. Here, the cohort of individuals may change their behavior in response to what they learned about the measured effects at previous time points of the longitudinal study. This constitutes a sample adaptation mapping from the sample and the fitted model parameters at time $t$ to the sample at time $t+1$. By checking whether this mapping is Lipschitz, %
researchers can use our results to guarantee valid inference from the effects measured at time $T$. (However, in this setting, unlike in machine learning, %
the sample adaptation mapping is out of the researcher's direct control and must be learned based on observed cohorts.)
As briefly touched upon in Section~\ref{sec:outline}, we consider the development of learning-theory-informed methods---whether in machine learning or otherwise---to be the greatest potential use case of the present work. 

A second avenue for future work is to obtain sharper theoretical guarantees for specific reciprocal learning paradigms.
In particular, the capabilities of active learning, bandits and semi-supervised learning to generalize can be investigated beyond the bounds derived in this paper. 
As our results require no distributional assumptions on the newly added data, tighter bounds appear within reach by making such assumptions, tailored to the particular paradigm under investigation. 
As a simple example, consider the setting in which, at each iteration $t$, a reciprocal learning algorithm replaces $m_t$ sample points by fresh draws from some law $\tilde{\mathbb P}$. In this case, $\hat{\mathbb P}_t$ consists of those observations retained from previous rounds together with newly
sampled observations from $\tilde{\mathbb P}$; in general, it is therefore not i.i.d.\ from
$\tilde{\mathbb P}$ nor from $\mathbb P$. While $\tilde{\mathbb P}$ is often unknown in practice, background knowledge
may justify assuming a specific distribution shift from $\mathbb P$ to $\tilde{\mathbb P}$---for instance covariate shift, target shift or conditional shift \citep[see][]{pmlr-v28-zhang13d}. 

Alternatively, the reciprocal learning algorithm in question may itself impose a kind of distribution shift, as we now consider. To illustrate the potential of adapting our results to such %
instances of reciprocal learning, 
consider the special case of boosting. Here, the algorithm changes only labels while keeping the covariates of the modified points fixed. Recall the simple example from Section~\ref{sec:intro}: gradient boosting’s pseudo-residuals correspond to the residuals $y- \hat y$. 
At iteration $t$, gradient boosting thus fits a model to the residuals $y - \hat y(\theta_{t-1}, x)$ from the previous iteration $t-1$. It is easy to see that in this case, we can tighten our bounds by replacing $\mathcal Z$'s diameter $\diamZ$ in Lemma~\ref{lemma:recipr-distortion} with the diameter $\diamY$ of the label space $\mathcal Y$. The following result spells out this idea. %

\begin{prop}[Improved Bounds for Boosting]\label{prop:boosting}
        Assume that reciprocal learning is non-greedy and modifies $m_{t+1}$ of the $n_t$ %
    sample points when transitioning from $\hat{\mathbb P}_t$ to $\hat{\mathbb P}_{t+1}$.  %
    Further, suppose
    that every modified point keeps its covariates fixed and only changes its labels (as in boosting, for example).
    Then Lemma~\ref{lemma:recipr-distortion} holds with $\diamZ$ replaced by
    $
    \diamY:=\sup_{y,y'\in\mathcal Y} \| y - y'\|_2 <\infty.
    $
    Hence, the generalization gap bound (Theorem~\ref{thm:gen-error-bound}),
    the excess risk bound (Theorem~\ref{thm:excess-risk-classic}), the anytime-valid generalization gap bound (Theorem~\ref{thm:gen-error-bound-anytime}), and the anytime-valid excess risk bound (Theorem~\ref{thm:excess-risk-classic-anytime}) can all be tightened
    by replacing %
    the term $\left(\frac{m}{n}\right)^{\frac{1}{p}} \diamZ$ with $\left(\frac{m}{n}\right)^{\frac{1}{p}}\diamY$. %

\end{prop}
Beyond this simplistic example, more sophisticated background knowledge on the shift from $\mathbb P$ to $\tilde{ \mathbb P}$ \citep[e.g., through invariant feature representations---see][]{pmlr-v28-muandet13,muandet2017kernel} could give rise to tighter bounds. However, this would require stronger assumptions than our general analysis. One possible middle ground between our analysis---which, as we have emphasized, does not require any assumptions regarding the distribution of the added data---and assuming the added data are drawn from an (at least partially known) probability distribution $\tilde{ \mathbb P}$ could lie in the field of imprecise probabilities \citep{walley1991statistical,augustin2014introduction,de2015imprecision}. Specifically, the recently introduced \say{imprecise domain generalization} framework allows learners to stay imprecise, by not requiring them to commit to a fixed $\mathbb{P}$, nor to a fixed $\tilde{ \mathbb P}$, nor to a mixture of both \citep{singh2024domain}. Choosing appropriate learners under this kind of second-order uncertainty can then be guided by decision-theoretic criteria along the lines of \citet{de2008strategy,antonucci2011decision,jansen2022quantifying,jansen2023statistical,jansen2023robust,jansen-gsd-24}. Notably, imprecise extensions of some of reciprocal learning's special cases like active learning \citep{nguyen2022measure,hullermeier2022quantification} or Bayesian optimization \citep{rodemann2021accounting,imprecise-BO} already exist---pointing to the possibility of an imprecise generalization of reciprocal learning.

Closely related to both our work and imprecise probability, \citet{caprio2024credal} derive generalization bounds that are robust to second-order uncertainty \citep[see also][]{caprio2024credal-bayesian}. They require the specification of sets of distributions, making them more general than classical learning theory, but still more specific than our analysis, which requires no assumption on the distribution of acquired data at all.   

We propose one final direction for future work: 
reducing the dependence of our results on the dimension $\dz$ of $\mathcal Z$. 
Similar concentration rates to those in Lemma~\ref{lemma-wasserstein-iid} could potentially be achieved independently of the dimension $\dz$ by considering the Kolmogorov--Smirnov distance instead of the Wasserstein distance and using integration by parts \citep[see][]{gaunt2023bounding}. %

\section{Acknowledgments and Disclosure of Funding}

We are indebted to Thomas Nagler (LMU Munich), Christoph Jansen (Lancaster University), Georg Schollmeyer (LMU Munich), Xiao-Li Meng (Harvard University), Thomas Augustin (LMU Munich) and Natesh Pillai (Harvard University) for helpful comments, stimulating discussions, and detailed feedback at various stages of this work. Furthermore, we wish to thank the participants of the second \textit{Workshop on Machine Learning under Weakly Structured Information in T\"ubingen} (\url{https://fm.ls/luwsi2025}) for their critical assessment of the ideas related to this article, as they were presented at the workshop. All remaining errors are solely our own. JR gratefully acknowledges funding by the federal German Academic Exchange Service (DAAD) supporting a research stay at Harvard University in 2024. This work was partially conducted while JB was at the Department of Statistics, Harvard University in Cambridge, USA. JB gratefully acknowledges partial financial support from the Australian-American Fulbright Commission and the Kinghorn Foundation.

\bibliography{bib}
\addcontentsline{toc}{section}{References}

\appendix

\section{Proofs}\label{app:proofs}

\subsection{Proof of Lemma~\ref{lemma:recipr-distortion}}

For ease of exposition, we restate Lemma~\ref{lemma:recipr-distortion} before proving it:\\ 

\noindent \textbf{Lemma (Reciprocal Distortion Bound)}
    {\itshape    Let $\hat{\mathbb P}_{T}$ be the empirical distribution of training data at iteration $T$ in reciprocal learning and $\hat{\mathbb P}_0$ the initial one. If the sample adaptation function $f_s$ is almost surely $L_s$-Lipschitz (Condition~\ref{cond:lipschitz-sample-adpat}), then

    \[    W_p(\hat{\mathbb P}_0, \hat{\mathbb P}_{T}) \leq  S_T(L_s) \left(\frac{m}{n}\right)^{\frac{1}{p}} \diamZ =: \beta_T, %
\]
        pointwise in $\hat{\mathbb P}_0$ and almost surely pathwise in $\hat{\mathbb P}_T$, where $\diamZ := \sup_{z, z' \in \mathcal Z} \| z - z \|_2 < \infty$ %
    is the diameter of $\mathcal{Z}$ and  %
}
    \[     S_T(L_s) = \begin{cases}
        \frac{L_s^{T} - 1}{L_s - 1} & \mathrm{if\ } L_s \ne 1,\\
        T & \mathrm{if\ } L_s = 1.
    \end{cases}
\] 

\begin{proof}
    Recall that $\hat{\mathbb P}_{t}$ is the empirical distribution of the training data at iteration $t$ of reciprocal learning and that $\hat{\mathbb P}_0$ describes the initial sample that is drawn \iid\ from $\mathbb{P}$. 

    Define 
    \begin{equation}\label{eq:defn-closed-loop-sample-adaptation}
        \Phi(P) := f_s(\hat\theta(P),P),
    \end{equation}
    where $\hat\theta(P) \in \arg\min_{\theta\in\Theta} \mathscr R(P,\theta)$ is the reciprocal learner's fitted parameter given the distribution $P \in \mathcal P (\mathcal Z)$. Let $\omega \in \Omega$ be the random seed that is used by the sample adaptation function $f_s$, and write $f_s(\,\cdot\,,\, \cdot\,)(\omega)$ for the realized sample adaptation function given the random seed $\omega$. Similarly define $\Phi(\,\cdot\,)(\omega) := f_s(\hat\theta(\,\cdot\,),\,\cdot\,)(\omega)$ to be the realized version of $\Phi$ given $\omega$. Condition~\ref{cond:lipschitz-sample-adpat} implies that
    \begin{equation}\label{eq:defn-Phi-Lips-cts}
        W_p \left[ \Phi (P)(\omega), \Phi (Q)(\omega') \right] \le L_s W_p \left[ P, Q \right],
    \end{equation}
    for all $P, Q \in \mathcal P(\mathcal Z)$ and almost all $\omega, \omega'$.
    (To be clear, the ``almost all'' quantifier on the random seeds $\omega$ and $\omega'$ is over the product distribution $\mathbb P_{\omega} \times \mathbb P_{\omega}$, where $\mathbb P_{\omega} \in \mathcal P(\Omega)$ is the probability distribution from which $f_s$ draws its random seed.)
    
    In the case of deterministic sample adaptation, consider the fixed sample path $\hat{\mathbb P}_0, \ldots \hat{\mathbb P}_T$ induced by the initial sample $\hat{\mathbb P}_0$. In the case of random $f_s$, %
    fix a sample path $\hat{\mathbb P}_{0}, \dots, \hat{\mathbb P}_{T}$ along which $\Phi$ is $L_s$-Lipschitz continuous (which is almost all sample paths by Condition~\ref{cond:lipschitz-sample-adpat}). More specifically, %
    for all $t \in \{1,\ldots,T\}$, choose a random seed $\omega_t$ so that $\omega = \omega_t$ and $\omega' = \omega_{t-1}$ satisfy Equation~\ref{eq:defn-Phi-Lips-cts} for all $P, Q \in \mathcal P(\mathcal Z)$ and all $t \in \{2,\ldots, T\}$.
    
    By the triangle inequality, we get
    \begin{equation}\label{eq:dist-triangle}
            W_p(\hat{\mathbb P}_0, \hat{\mathbb P}_{T}) \leq \sum_{t=0}^{T-1} W_p(\hat{\mathbb P}_{t}, \hat{\mathbb P}_{t+1}).
    \end{equation}
    By Equation~\ref{eq:defn-Phi-Lips-cts}%
    , we further have that
    \begin{equation*}
        W_p(\hat{\mathbb P}_{t+1}, \hat{\mathbb P}_{t}) \leq L_s W_p(\hat{\mathbb P}_{t}, \hat{\mathbb P}_{t-1}),
    \end{equation*}
    and thus 
    \begin{equation*}
        W_p(\hat{\mathbb P}_{t+1}, \hat{\mathbb P}_{t}) \leq L_s  W_p(\hat{\mathbb P}_{t}, \hat{\mathbb P}_{t-1}) \leq L_s^2  W_p(\hat{\mathbb P}_{t-1}, \hat{\mathbb P}_{t-2}) \leq \ldots \leq L_s^{t} W_p(\hat{\mathbb P}_{1}, \hat{\mathbb P}_0).
    \end{equation*}
    Summing both sides over $t = 0, \dots, T-1$ yields
    \begin{equation}\label{eq:geomseries}
        \sum_{t=0}^{T-1} W_p(\hat{\mathbb P}_{t+1}, \hat{\mathbb P}_{t}) \leq \sum_{t=0}^{T-1} L_s^{t}  W_p(\hat{\mathbb P}_{1}, \hat{\mathbb P}_0) = S_T(L_s) W_p(\hat{\mathbb P}_{1}, \hat{\mathbb P}_0).
    \end{equation}

    Observe that $W_p(\hat{\mathbb P}_{1}, \hat{\mathbb P}_0) = W_p(f_s(\hat{\theta}_0, \hat{\mathbb P}_0), \hat{\mathbb P}_0)$. Since $f_s$ changes $m$ units in the initial sample and $\mathcal{Z}$ is bounded by $\diamZ$, we can further bound
    \begin{equation}\label{eq:diameter}
        W_p(f_s(\hat{\theta}_0, \hat{\mathbb P}_0), \hat{\mathbb P}_0) \leq \left(\frac{m }{n}\right)^{\frac{1}{p}} \diamZ,
    \end{equation}
    by the definition of the $p$-Wasserstein distance. %
    By combining Equations~\ref{eq:dist-triangle}, \ref{eq:geomseries} and~\ref{eq:diameter}, the desired result follows immediately in the case of deterministic $f_s$. If $f_s$ is random, the same chain of inequalities holds for every sample path along which $\Phi$ is $L_s$-Lipschitz%
    , since the above reasoning is a deterministic statement about a fixed path. By Condition~\ref{cond:lipschitz-sample-adpat}, these are all paths outside a null set, so the desired bound holds almost surely pathwise.
    More exactly, the set 
    \[\{ (\omega_1, \ldots, \omega_T) : W_p \left[ \Phi (P)(\omega_t), \Phi (Q)(\omega_{t-1}) \right] \le L_s W_p \left[ P, Q \right] \mathrm{\ for\ all\ } P, Q \in \mathcal P(\mathcal Z) \},\]
    
    has probability one by Condition~\ref{cond:lipschitz-sample-adpat}. Since we have proved the result for every $(\omega_1,\ldots,\omega_T)$ in this set, the result holds almost surely. 
\end{proof}

\subsection{Proof of Theorem~\ref{thm:gen-error-bound}}

We first restate Theorem~\ref{thm:gen-error-bound} for readability:\\

\noindent \textbf{Theorem (Generalization Gap)}
{\itshape    Suppose the loss is $L_{\ell}$-Lipschitz in the data (Condition~\ref{ass-cont-diff-feat-param-Lips-data}) and reciprocal learning converges to $(\hat{\theta}_c, \hat{\mathbb P}_c)$ (Condition~\ref{cond:conv-reciprocal-learning}) with $L_s$-Lipschitz sample adaptation (Condition~\ref{cond:lipschitz-sample-adpat}) where $L_s < 1$. Then

    \[
            \mathscr R(\mathbb P, \hat{\theta}_c) \leq  \mathscr R(\hat{\mathbb P}_c, \hat{\theta}_c) + L_\ell \left(\frac{\log \left(C_a / \delta\right)}{C_{b} n}\right)^{p / d_z} + \frac{ L_\ell \left(\frac{m }{n}\right)^{\frac{1}{p}} \diamZ}{1 - L_s} ,

    \]
        with probability at least $1-\delta$%
    , where $C_a$ and $C_b$ are constants (from Lemma~\ref{lemma-wasserstein-iid}) which depend on $p$, $\dz$ and $\diamZ$. %
}\\

\begin{proof}
    Let $\Phi(P) := f_s(\hat\theta(P),P)$ be the map defined in Equation~\ref{eq:defn-closed-loop-sample-adaptation}. For now, fix a sample path $\hat{\mathbb P}_0, \hat{\mathbb P}_1, \ldots$ along which $\Phi$ is $L_s$-Lipschitz continuous. (More exactly, follow the strategy in the proof of Lemma~\ref{lemma:recipr-distortion} that fixes random seeds $\omega_t$ at each iteration $t\ge 1$ so that $\omega = \omega_t$ and $\omega' = \omega_{t-1}$ satisfy Equation~\ref{eq:defn-Phi-Lips-cts} for all $P, Q \in \mathcal P(\mathcal Z)$.) %
    We have---via the Lipschitz continuity of the loss in the data (Condition~\ref{ass-cont-diff-feat-param-Lips-data})---by the Kantorovich--Rubinstein lemma (Lemma~\ref{lemma:kant-rubin}) that
\begin{equation}\label{eq:proof:thm-gen-error-bound-KRLemma}
    | \mathscr R(\mathbb P, \hat{\theta}_c) - \mathscr R(\hat{\mathbb P}_c, \hat{\theta}_c)  | \le L_\ell W_1(\mathbb P, \hat{\mathbb P}_c),
\end{equation}
where $L_\ell$ is the Lipschitz constant of $\ell$.
We further have $W_1 (\mathbb P, \hat{\mathbb P}_c) \leq W_p(\mathbb P, \hat{\mathbb P}_c)$  by monotonicity in $p$ of the $p$-Wasserstein distance. %
In other words, we can bound the risk difference $\mathscr R(\mathbb P, \hat{\theta}_c) - \mathscr R(\hat{\mathbb P}_c, \hat{\theta}_c)$ via the Wasserstein distance $W_p(\mathbb P, \hat{\mathbb P}_c)$ between its respective distributions. 

We thus have to bound $W_p(\mathbb P, \hat{\mathbb P}_c)$. 
It holds by Lemma~\ref{lemma:recipr-distortion} that 
    \[W_p(\hat{\mathbb P}_0, \hat{\mathbb P}_T) \leq \frac{1}{1 - L_s}  \left(\frac{m }{n}\right)^{\frac{1}{p}} \diamZ,\]
for any $T$,
and thus also for $\hat{\mathbb P}_c$. %
Further, Lemma~\ref{lemma:wasserstein-conv} implies
$
    W_p(\mathbb P, \hat{\mathbb P}_0) \leq \left(\frac{\log \left(C_a / \delta\right)}{C_{b} n}\right)^{p / d_z}
$
with probability over $\hat{\mathbb P}_{0}$ of at least $1 - \delta$. 
This gives
\begin{equation*}    
 W_p(\mathbb P, \hat{\mathbb P}_c) \leq  W_p(\mathbb P, \hat{\mathbb P}_0) + W_p(\hat{\mathbb P}_0, \hat{\mathbb P}_c)  %
    \leq \left(\frac{\log \left(C_a / \delta\right)}{C_{b} n}\right)^{p / d_z} +    \frac{1}{1 - L_s}  \left(\frac{m }{n}\right)^{\frac{1}{p}} \diamZ,    
\end{equation*}
    with probability over $\hat{\mathbb P}_{0}$ of at least $1 - \delta$. %
This---together with Equation~\ref{eq:proof:thm-gen-error-bound-KRLemma}---gives 
\begin{equation*}
\mathscr R(\mathbb P, \hat{\theta}_c) \leq  \mathscr R(\hat{\mathbb P}_c, \hat{\theta}_c) + L_\ell \left(\frac{\log \left(C_a / \delta\right)}{C_{b} n}\right)^{p / d_z} +    \frac{L_\ell}{1 - L_s}  \left(\frac{m }{n}\right)^{\frac{1}{p}} \diamZ, 
\end{equation*}
with probability over $\hat{\mathbb P}_{0}$ of at least $1 - \delta$. This proves the result for a fixed sample path along which $\Phi$ is $L_s$-Lipschitz continuous (more specifically, for fixed random seeds $\omega_t$ at each iteration $t = 1, 2, \ldots$ for which $\omega = \omega_t$ and $\omega' = \omega_{t-1}$ satisfy Equation~\ref{eq:defn-Phi-Lips-cts}). With the same reasoning as in the proof of Lemma~\ref{lemma:recipr-distortion}, the result must therefore also hold almost surely pathwise (i.e., almost surely over all the reciprocal paths).
\end{proof}

\subsection{Proof of Theorem~\ref{thm:gen-error-bound-data-dependent}}

For the reader’s convenience, we begin by restating Theorem~\ref{thm:gen-error-bound-data-dependent}:\\ 

\noindent \textbf{Theorem (Data-Dependent Excess Risk Bound)}
{\itshape    Suppose the loss is $L_{\ell}$-Lipschitz in the data (Condition~\ref{ass-cont-diff-feat-param-Lips-data}) and reciprocal learning converges to $(\hat{\theta}_c, \hat{\mathbb P}_c)$ (Condition~\ref{cond:conv-reciprocal-learning}). Then the excess risk $\mathscr R({\mathbb P}, \hat{\theta}_c) - \mathscr R_\Theta $ is upper bounded by

    \begin{equation*}
                \mathscr R(\hat{\mathbb P}_0, \hat{\theta}_c) - \mathscr R(\hat{\mathbb P}_0, \hat{\theta}_0) + L_\ell \left(\frac{\log \left(C_a / \delta\right)}{C_{b} n}\right)^{p / d_z}  +
        \frac{24}{\sqrt n} \mathfrak C_{L_2}(\mathcal L_0) + L_\ell \diamZ \sqrt{\frac{\log(1/\delta)}{2n}},

    \end{equation*}
        with probability at least $1-2\delta$.
}\\

\begin{proof}
This proof relies on Lemmas \ref{lemma:excess-risk-classic-1} and \ref{lemma:excess-risk-classic-3} from the proof of Theorem~\ref{thm:excess-risk-classic}. Namely, given Condition~\ref{ass-cont-diff-feat-param-Lips-data}, we have %
$
    \mathscr R({\mathbb P}, \hat{\theta}_c) - \mathscr R(\hat{\mathbb P}_0, \hat{\theta}_c) \leq   L_\ell \left(\frac{\log \left(C_a / \delta\right)}{C_{b} n}\right)^{p / d_z}
$ with probability over $\hat{\mathbb P}_{0}$ of at least $1 - \delta$. (This is Lemma~\ref{lemma:excess-risk-classic-1}.) 
Furthermore, %
\begin{equation*}%
    \mathscr R(\hat{\mathbb P}_0, \hat{\theta}_0) - \mathscr R_\Theta \leq \frac{24}{\sqrt n} \mathfrak{C}_{L_2}(\mathcal L_0) + L_\ell \diamZ \sqrt{\frac{\log(1/\delta)}{2n}},
\end{equation*}
with probability over $\hat{\mathbb P}_{0}$ of at least $1 - \delta$. (This is Lemma~\ref{lemma:excess-risk-classic-3}.) Appendix~\ref{proof:excess-risk-classic} proves that both these statements hold almost surely over the randomness in the sample adaptation function. 
The claim then follows by the fact that
\begin{equation*}
    \mathscr R({\mathbb P}, \hat{\theta}_c) - \mathscr R_\Theta = \left(\mathscr R({\mathbb P}, \hat{\theta}_c) - \mathscr R(\hat{\mathbb P}_0, \hat{\theta}_c) \right) + \left( \mathscr R(\hat{\mathbb P}_0, \hat{\theta}_c) - \mathscr R(\hat{\mathbb P}_0, \hat{\theta}_0) \right)
    + \left( \mathscr R(\hat{\mathbb P}_0, \hat{\theta}_0) - \mathscr R_\Theta \right),
\end{equation*}
as well as a union bound argument.
\end{proof}

\subsection{Proof of Corollary~\ref{cor:gen-error-bound-data-dependent}}\label{proof:gen-error-bound-data-dependent}

We repeat the corollary:\\

\noindent \textbf{Corollary (Excess Risk Bound in Terms of Hypothesis Space Complexity)}
{\itshape    Assume that the loss function $\ell$ takes the form
    \[\ell(\theta, x, y) = \tilde \ell \bigl( f_\theta(x), y \bigr) + r(\theta),\]
    where $r : \Theta \to \mathbb R$ is an arbitrary regularizer and $\tilde \ell : \mathcal Y \times \mathcal Y \to \mathbb R$ is the unpenalized loss function. (We place no restrictions on the form of $r$; in particular, an unpenalized loss $\ell$ can be handled by taking $r(\theta) = 0$.) Further suppose that $\tilde \ell (\,\cdot\,, y')$ is $L_\ell$-Lipschitz for all $y' \in \mathcal Y$. Then we may replace $\mathfrak C_{L_2}(\mathcal L_0)$ with $L_{\ell} \mathfrak C_{L_2}(\mathcal F)$ in each of the excess risk bounds---that is, in Theorems~\ref{thm:gen-error-bound-data-dependent},~\ref{thm:excess-risk-classic},~\ref{thm:excess-risk-classic-anytime} and~\ref{thm:data-dependent-anytime-excess}.
}\\ 

\begin{proof}
    We only need to modify Part~\ref{proof:excess-risk-classic-3} of the proof of Theorem~\ref{thm:excess-risk-classic}, because this is the only place that $\mathfrak C_{L_2}(\mathcal L_0)$ appears. (Note that the relevant parts of the proofs of Theorems~\ref{thm:gen-error-bound-data-dependent},~\ref{thm:excess-risk-classic-anytime} and~\ref{thm:data-dependent-anytime-excess} all rely on Part~\ref{proof:excess-risk-classic-3} of the proof of Theorem~\ref{thm:excess-risk-classic}, so modifying this part will automatically prove the modified versions of Theorems~\ref{thm:gen-error-bound-data-dependent},~\ref{thm:excess-risk-classic-anytime} and~\ref{thm:data-dependent-anytime-excess} too.) Let $\tilde \ell_\theta$ be the map $z \mapsto \tilde \ell(f_\theta(x), y)$ (where $z = (x,y)$) and define $\tilde{\mathcal L} := \{\tilde \ell_\theta : \theta \in \Theta\}$ to be the class of unpenalized loss functions. We can rewrite $G = \sup_{\theta \in \Theta} \mathscr R(\hat{\mathbb P}_0, \theta) - \mathscr R(\mathbb P, \theta)$ from Part~\ref{proof:excess-risk-classic-3} of the proof of Theorem~\ref{thm:excess-risk-classic} as
    \[ G = \sup_{\tilde \ell_\theta \in \tilde{\mathcal L}} \left[ \mathbb E_{Z \sim \hat{\mathbb P}_0} [\tilde \ell_{\theta} (Z)] - \mathbb E_{Z \sim \mathbb P} [ \tilde \ell_\theta(Z)]\right].\]
    Thus, analogously to Equation~\ref{eq:rademacher}, we have $\mathbb E[G] \le 2 \mathfrak R_n (\tilde{\mathcal L})$.

    It remains to bound $\mathfrak R_n (\tilde{\mathcal L})$.
    Fix $Q \in \mathcal P(\mathcal Z)$ and let $Q_{\mathcal X}$ denote its marginal on $\mathcal X$. By the fact that $\tilde \ell(y, y')$ is $L_\ell$-Lipschitz in $y$ for all $y'$, we get that, for all $\theta, \theta' \in \Theta$,
    \begin{align*}
        \left\| \tilde \ell_\theta - \tilde \ell_{\theta'} \right\|^2_{L_2(Q)} &= \int_{\mathcal Z} \left| \tilde \ell \bigl(f_\theta(x), y\bigr) - \tilde \ell \bigl(f_{\theta'}(x), y \bigr) \right|^2 \mathrm{d} Q(x,y) \\
        & \leq L_\ell^2 \int_{\mathcal X} \left\| f_\theta(x) - f_{\theta'}(x) \right\|_2^2 \mathrm{d} Q_{\mathcal X}(x) \\
        & = L_{\ell^2} \| f_\theta - f_{\theta'} \|_{L_2(Q_\mathcal X)}^2.
    \end{align*}
    Thus, every $\varepsilon / L_\ell$-cover of $\mathcal F$ with respect to $\|\cdot\|_{L_2(Q_{\mathcal X})}$ induces an $\varepsilon$-cover of $\tilde{\mathcal L}$ with respect to $\|\cdot\|_{L_2(Q)}$. That is, $\mathcal N \bigl( \tilde{\mathcal L}, \|\cdot\|_{L_2(Q)}, \varepsilon \bigr) \leq \mathcal N \bigl( \mathcal F, \|\cdot\|_{L_2(Q_{\mathcal X})}, \varepsilon / L_\ell \bigr)$. Substituting $u = \varepsilon / L_\ell$ in the covering entropy integral and taking the supremum over $Q$, Dudley's theorem \citep{dudley1987universal} therefore yields
    \begin{equation*}%
        \Re_n(\tilde{\mathcal L})  \leq  \frac{12}{\sqrt{n}} \sup_{Q \in \mathcal P(\mathcal Z)} \int_{0}^{\infty} \sqrt{\log \mathcal N \bigl( \tilde{\mathcal L}, \|\cdot\|_{L_2(Q)}, \varepsilon \bigr)}  \mathrm{d} \varepsilon  \leq  \frac{12 L_\ell}{\sqrt{n}}  \mathfrak{C}_{L_2}(\mathcal{F}).
    \end{equation*}
    Hence,
    \[ \mathbb E[G] \le \frac{24 L_\ell}{\sqrt{n}} \mathfrak{C}_{L_2}(\mathcal{F}).\]
    The remainder of Part~\ref{proof:excess-risk-classic-3} of the proof of Theorem~\ref{thm:excess-risk-classic} remains unchanged, so that we can conclude
    \begin{equation}\label{eq:cor-new-bound}
        \mathscr R(\hat{\mathbb P}_0, \hat{\theta}_0) - \mathscr R_\Theta 
         \leq  
        \frac{24 L_\ell}{\sqrt{n}} \mathfrak{C}_{L_2}(\mathcal{F})
        + L_\ell \diamZ \sqrt{\frac{\log(1/\delta)}{2n}},
    \end{equation}
with probability over $\hat{\mathbb P}_{0}$ of at least $1 - \delta$. Each of the proofs for Theorems~\ref{thm:gen-error-bound-data-dependent},~\ref{thm:excess-risk-classic},~\ref{thm:excess-risk-classic-anytime} and~\ref{thm:data-dependent-anytime-excess} uses Part~\ref{proof:excess-risk-classic-3} of the proof of Theorem~\ref{thm:excess-risk-classic} only through the bound it establishes on $\mathscr R(\hat{\mathbb P}_0, \hat{\theta}_0) - \mathscr R_\Theta$. Therefore, substituting Equation~\ref{eq:cor-new-bound} for that bound in each of those proofs, and leaving the remainder of each proof unchanged, proves the corollary.
\end{proof}

\subsection{Proof of Theorem~\ref{thm:excess-risk-classic}}\label{proof:excess-risk-classic}

We begin with a restatement of Theorem~\ref{thm:excess-risk-classic} for clarity:\\ 

\noindent \textbf{Theorem (Excess Risk Bound)}
{\itshape   Suppose $\Theta$ is closed and convex. Assume the loss is continuously differentiable and $L_{\ell}$-Lipschitz (Condition~\ref{ass-cont-diff-feat-param}) as well as strongly convex (Condition~\ref{ass-str-convex}). If reciprocal learning converges to $(\hat{\theta}_c, \hat{\mathbb P}_c)$ (Condition~\ref{cond:conv-reciprocal-learning}) with $L_s$-Lipschitz sample adaptation (Condition~\ref{cond:lipschitz-sample-adpat}), where $L_s < 1$, then its excess risk
    $ \mathscr R({\mathbb P}, \hat{\theta}_c) - \mathscr R_\Theta $ is upper bounded by
    \[
        L_\ell \left(\frac{\log \left(C_a / \delta\right)}{C_{b} n}\right)^{p / d_z} +  \frac{L_\ell  (\kappa/\gamma) \left(\frac{m }{n}\right)^{\frac{1}{p}} \diamZ}{ 1 - L_s } +
        \frac{24}{\sqrt n} \mathfrak C_{L_2}(\mathcal L_0) + L_\ell \diamZ \sqrt{\frac{\log(1/\delta)}{2n}},
    \]
    with probability at least $1-2\delta$.
}\\ 

\begin{proof}
The structure of the proof is given by the following three lemmas:
\begin{enumerate}[label=(\Alph*)]
    \item\label{lemma:excess-risk-classic-1} We first show via the Kantorovich--Rubinstein lemma that
$
    \mathscr R({\mathbb P}, \hat{\theta}_c) - \mathscr R(\hat{\mathbb P}_0, \hat{\theta}_c) \leq   L_\ell \left(\frac{\log \left(C_a / \delta\right)}{C_{b} n}\right)^{p / d_z} 
$
with probability over $\hat{\mathbb P}_{0}$ of at least $1 - \delta$. 
\item\label{lemma:excess-risk-classic-2} We then show  
$\mathscr R(\hat{\mathbb P}_0, \hat{\theta}_c) - \mathscr R(\hat{\mathbb P}_0, \hat{\theta}_0)   \leq \frac{L_\ell  (\kappa/\gamma) \left(\frac{m }{n}\right)^{\frac{1}{p}} \diamZ}{1 - L_s}$ almost surely pathwise.

\item\label{lemma:excess-risk-classic-3} We prove via a standard symmetrization argument that $\mathscr R(\hat{\mathbb P}_0, \hat{\theta}_0) - \mathscr R_\Theta$ is upper bounded by $\frac{24}{\sqrt n} \mathfrak{C}_{L_2}(\mathcal L_0) + L_\ell \diamZ \sqrt{\log(1/\delta) / (2n)}$ with probability over $\hat{\mathbb P}_{0}$ of at least $1 - \delta$.

\end{enumerate}

\noindent The theorem then follows by a union bound over these lemmas.
To get some intuition for these lemmas, note that \ref{lemma:excess-risk-classic-1} relates the generalization error of reciprocal learning's output $\hat \theta_c$ to the risk on the initial \iid~sample $\hat{\mathbb P}_0$; \ref{lemma:excess-risk-classic-2} compares this risk to the one of the initial one-shot empirical risk minimizer $\hat{\theta}_0$; and \ref{lemma:excess-risk-classic-3} utilizes that the difference between $\hat{\theta}_0$'s risk and the Bayes risk $\mathscr R_\Theta$ constitutes a classical empirical process \citep[see, e.g.,][]{vaart1997weak}.  

We now prove each of the three lemmas in turn:

\begin{enumerate}[label=(\Alph*)]
\item\label{proof:excess-risk-classic-1} %
By analogous reasoning to the proof of Theorem~\ref{thm:gen-error-bound},
    \[ |\mathscr R(\mathbb P, \hat{\theta}_c) - \mathscr R(\hat{\mathbb P}_0, \hat{\theta}_c)  | \le L_\ell W_1(\mathbb P, \hat{\mathbb P}_0) \le L_{\ell} W_p(\mathbb P, \hat{\mathbb P}_0) \le L_\ell \left(\frac{\log \left(C_a / \delta\right)}{C_{b} n}\right)^{p / d_z},\]
where the first inequality follows from Condition~\ref{ass-cont-diff-feat-param-Lips-data} by the Kantorovich--Rubinstein lemma (Lemma~\ref{lemma:kant-rubin}), the second is implied by monotonicity of $p$, and the third holds with probability over $\hat{\mathbb P}_{0}$ of at least $1 - \delta$ by Lemma~\ref{lemma:wasserstein-conv}.
\item\label{proof:excess-risk-classic-2}
Let $\Phi(P) := f_s(\hat\theta(P),P)$ be the map defined in Equation~\ref{eq:defn-closed-loop-sample-adaptation}. Fix a reciprocal path along which $\Phi$ is $L_s$-Lipschitz continuous. As in the previous proofs, we will relax this assumption at the end of the proof of this lemma. 

We will first bound $\|\hat{\theta}_0 - \hat{\theta}_T \|_2$  and then let $T \rightarrow \infty$ to bound $\|\hat{\theta}_0 - \hat{\theta}_c \|_2$, which will then allow us to bound $\mathscr R(\hat{\mathbb P}_0, \hat{\theta}_c) - \mathscr R(\hat{\mathbb P}_0, \hat{\theta}_0)$ via Lipschitz continuity of the loss in $\theta$ (Condition~\ref{ass-cont-diff-feat-param-cts-diff-param}). %

On our way to bounding $\|\hat \theta_0 - \hat \theta_T \|_2$, we begin by showing that the ERM map %
$\hat \theta (P) := \argmin_{\theta\in\Theta}\mathscr R(P,\theta)$
is well defined and Lipschitz continuous from $(\mathcal P(\mathcal Z),W_p)$ to
$(\Theta,\|\cdot\|_2)$, with Lipschitz constant at most $L_a := \kappa/\gamma$. Conditions~\ref{ass-cont-diff-feat-param-cts-diff-param}
and~\ref{ass-str-convex} imply that, for every $P\in\mathcal P(\mathcal Z)$,
the map $\theta\mapsto\mathscr R(P,\theta)$ is
differentiable and $\gamma$-strongly convex, with
\begin{equation}\label{eq:grad-exchange}
\nabla_\theta\mathscr R(P,\theta)
=
\int_{\mathcal Z}\nabla_\theta\ell(\theta,z)\mathrm dP(z).
\end{equation}
(We can swap the order of integration and differentiation by the Leibniz integral rule since $\mathcal Z$ is bounded and hence $\nabla_\theta\ell(\theta,z)$ is integrable.) Thus, because $\Theta$ is closed and convex, and $\theta\mapsto\mathscr R(P,\theta)$ is $\gamma$-strongly convex, the ERM map $\hat \theta$ is well defined---i.e., there exists a unique minimizer of $\mathscr R(P,\theta)$ for all $P \in \mathcal P(\mathcal Z)$.

To see that the ERM map $\hat \theta$ is $L_a$-Lipschitz, note that $\nabla_\theta\mathscr R(P,\,\cdot\,)$ is $\gamma$-strongly monotonic. That is, 
\[\langle \nabla_\theta \mathscr R(P,\theta) - \nabla_\theta \mathscr R(P,\theta'), \theta - \theta'\rangle \geq \gamma \|\theta - \theta'\|_2^2,\]
for all $\theta, \theta' \in \Theta$. To see this, observe that, for all $\theta,\theta' \in \Theta$,
\[0 \ge \left\langle \nabla_\theta \mathscr R(P,\theta), \theta'-\theta \right\rangle + \left\langle \nabla_\theta \mathscr R(P,\theta'), \theta-\theta' \right\rangle + \gamma\|\theta-\theta'\|_2^2,\]
by adding the following two equations together (which themselves come from the $\gamma$-strong convexity of the map $\theta\mapsto\mathscr R(P,\theta)$):
\begin{align*}
    \mathscr R(P,\theta') &\ge \mathscr R(P,\theta) + \nabla_\theta \mathscr R(P,\theta)^{\top} (\theta'-\theta) + \frac{\gamma}{2}\|\theta'-\theta\|_2^2, \\
    \mathscr R(P,\theta) &\ge \mathscr R(P,\theta') + \nabla_\theta \mathscr R(P,\theta')^{\top} (\theta-\theta') + \frac{\gamma}{2}\|\theta-\theta'\|_2^2.
\end{align*}

Now, fix $P,Q\in\mathcal P(\mathcal Z)$ and write
$\theta_P:=\hat \theta(P)$ and $\theta_Q:=\hat \theta(Q)$. For a differentiable convex
function $f$ on a convex set, a minimizer $\theta^\star$ satisfies
\begin{equation}\label{eq:first-order-optimality-condition}
    \langle\nabla f(\theta^\star),\theta-\theta^\star\rangle\geq0
\qquad\text{for every }\theta\in\Theta.
\end{equation}

Combining the above results, we have that
\begin{align}
    \gamma\|\theta_P-\theta_Q\|_2^2 &\leq \left\langle \nabla_\theta\mathscr R(P,\theta_P) - \nabla_\theta\mathscr R(P,\theta_Q), \theta_P - \theta_Q \right\rangle \notag\\
    &\leq \left\langle \nabla_\theta\mathscr R(Q,\theta_Q) - \nabla_\theta\mathscr R(P,\theta_Q), \theta_P - \theta_Q \right\rangle \notag\\
    &\leq \left\| \nabla_\theta\mathscr R(P,\theta_Q) - \nabla_\theta\mathscr R(Q,\theta_Q) \right\|_2 \|\theta_P - \theta_Q\|_2, \label{eq:erm-strong-conv}
\end{align}
where the first inequality uses $\gamma$-strong monotonicity of $\nabla_\theta\mathscr R(P,\,\cdot\,)$; the second uses the fact that $\left\langle \nabla_\theta\mathscr R(P,\theta_P), \theta_P - \theta_Q \right\rangle \le 0 \le \left\langle \nabla_\theta\mathscr R(Q,\theta_Q), \theta_P - \theta_Q \right\rangle $ by Equation~\ref{eq:first-order-optimality-condition}; and the last follows by the Cauchy--Schwarz inequality.

To bound the right hand side of Equation~\ref{eq:erm-strong-conv}, let $\|u\|_2=1$ and define
\[
h_u(z):=\langle\nabla_\theta\ell(\theta_Q,z),u\rangle .
\]
By Condition~\ref{ass-cont-diff-feat-param-cts-diff-param} and the Cauchy--Schwarz inequality, $h_u$ is $\kappa$-Lipschitz. Equation~\ref{eq:grad-exchange} and the Kantorovich--Rubinstein lemma (Lemma~\ref{lemma:kant-rubin}) therefore yield
\[\left| \left\langle \nabla_\theta\mathscr R(P,\theta_Q) - \nabla_\theta\mathscr R(Q,\theta_Q), u \right\rangle \right| \leq \kappa W_1(P,Q) \leq \kappa W_p(P,Q).\]
Using the Cauchy--Schwarz inequality, taking the supremum over all unit vectors $u$ gives
\begin{equation}\label{eq:erm-map-lipschitz}
    \|\hat \theta(P)-\hat \theta(Q)\|_2
    \leq L_aW_p(P,Q),
\end{equation}
Hence, the ERM map $\hat \theta$ is $L_a$-Lipschitz.

By Definition~\ref{def:basic-reciprocal-learning}, $\hat{\theta}_t = \hat \theta(\hat{\mathbb P}_t)$ for every $t$. Combining Equation~\ref{eq:erm-map-lipschitz} with Lemma~\ref{lemma:recipr-distortion} therefore gives
\begin{equation}\label{eq:theta-distortion}
    \|\hat{\theta}_0 - \hat{\theta}_{T}\|_2 = \|\hat \theta(\hat{\mathbb P}_0) - \hat \theta(\hat{\mathbb P}_{T})\|_2 \leq L_a W_p(\hat{\mathbb P}_0, \hat{\mathbb P}_{T}) \leq L_a \frac{L_s^{T} - 1}{L_s - 1} \left(\frac{m}{n}\right)^{\frac{1}{p}} \diamZ,
\end{equation}
for every $T$. By Condition~\ref{cond:conv-reciprocal-learning} and the Lipschitz continuity of $\hat \theta(\,\cdot\,)$, we further have $\hat{\theta}_c = \hat \theta(\hat{\mathbb P}_c)$, so that letting $T \rightarrow \infty$ and using $L_s < 1$ yields
\begin{equation*}%
    \|\hat{\theta}_0 - \hat{\theta}_{c}\|_2 \leq \frac{L_a \left(\frac{m}{n}\right)^{\frac{1}{p}} \diamZ}{1 - L_s}.
\end{equation*}
Since the loss is $L_\ell$-Lipschitz continuous in the parameters (Condition~\ref{ass-cont-diff-feat-param-cts-diff-param}), so is $\mathscr R(\hat{\mathbb P}_0, \,\cdot\,)$. Hence, we obtain
\begin{equation*}
    \mathscr R(\hat{\mathbb P}_0, \hat{\theta}_c) - \mathscr R(\hat{\mathbb P}_0, \hat{\theta}_0)  \leq L_\ell \|\hat{\theta}_0 - \hat{\theta}_{c}\|_2 \leq \frac{L_\ell L_a \left(\frac{m}{n}\right)^{\frac{1}{p}} \diamZ}{1 - L_s }.
\end{equation*}
This proves Lemma~\ref{lemma:excess-risk-classic-2} for any fixed reciprocal path along which $\Phi$ is $L_s$-Lipschitz continuous. With the same reasoning as in the proof of Lemma~\ref{lemma:recipr-distortion}, this means that Lemma~\ref{lemma:excess-risk-classic-2} must also hold almost surely over reciprocal paths.

\item\label{proof:excess-risk-classic-3} 
Fix $\eta > 0$ and pick $\theta_\eta \in \Theta$ with $\mathscr R(\mathbb P, \theta_\eta) \leq \mathscr R_\Theta + \eta$. (This is possible because $\mathscr R_\Theta$ is defined as an infimum.) Since $\hat \theta_0$ minimizes $\mathscr R(\hat{\mathbb P}_0, \,\cdot\,)$,
\begin{equation*}
    \mathscr R(\hat{\mathbb P}_0, \hat{\theta}_0) - \mathscr R_\Theta  \leq  \mathscr R(\hat{\mathbb P}_0, \theta_\eta) - \mathscr R(\mathbb P, \theta_\eta) + \eta  \leq \sup_{\theta \in \Theta} \bigl[ \mathscr R(\hat{\mathbb P}_0, \theta) - \mathscr R(\mathbb P, \theta) \bigl] +\, \eta. 
\end{equation*}
Defining $G%
= \sup_{\theta \in \Theta} \mathscr R(\hat{\mathbb P}_0, \theta) - \mathscr R(\mathbb P, \theta)$ (which is a function of the initial sample $Z_1, \ldots, Z_n \iidsim \mathbb P$ and hence is random) and taking $\eta \to 0^{+}$, we have 
\begin{equation}\label{eq:excess-risk-emp-process}
    \mathscr R(\hat{\mathbb P}_0, \hat{\theta}_0) - \mathscr R_\Theta  \leq G. %
\end{equation}
It remains to bound $G$. %
Write $\ell_\theta$ for the centered loss function $z \mapsto \ell(\theta, z) - \ell (\theta, z_0)$, so that $\mathcal L_0 = \{ \ell_\theta : \theta \in \Theta\}$. We can write 
\[ G = \sup_{\ell_\theta \in \mathcal L_0} \left[ \mathbb E_{Z \sim \hat{\mathbb P}_0} [\ell_{\theta} (Z)] - \mathbb E_{Z \sim \mathbb P} [ \ell_\theta(Z)]\right].\]
Then, by Condition~\ref{ass-cont-diff-feat-param-Lips-data}, $| \ell_\theta(z) -  \ell_\theta(z')| \leq L_\ell \|z - z'\|_2 \leq L_\ell \diamZ$ for all $\theta \in \Theta$ and all $z, z' \in \mathcal Z$. Replacing a single observation $Z_i$ in the initial sample $\hat{\mathbb P}_0$ therefore changes $G$ by at most $L_\ell \diamZ / n$, so that McDiarmid's inequality \citep{mcdiarmid1989method} gives
\begin{equation}\label{eq:mcdiarmid}
    G  \leq  \mathbb E[G] + L_\ell \diamZ \sqrt{\frac{\log(1/\delta)}{2n}},
\end{equation}
with probability at least $1 - \delta$. %

Now we bound $\mathbb E [G]$. %
Let
\begin{equation*}
    \Re_{n}(\mathscr{G}) := \mathbb{E}\left[\sup _{g \in \mathscr{G}} \frac{1}{n} \sum_{i=1}^{n} \mathcal{E}_{i} g \left(Z_{i}\right)\right]
\end{equation*}
denote the expected Rademacher average of a class $\mathscr G$ of real-valued functions on $\mathcal Z$. Here, the expectation is over both $Z_1, \ldots, Z_n \iidsim \mathbb P$ and $\mathcal{E}_1, \dots, \mathcal{E}_n$, which are \iid\ Rademacher random variables independent of $Z_1, \dots, Z_n$ \citep[see, e.g.,][]{bousquet2002stability,von2011statistical}. The standard symmetrization argument \citep[see, e.g., Chapter 26 of][]{shalev2014understanding}, applied to $\mathscr G = \mathcal L_0$, gives
\begin{equation}\label{eq:rademacher}
    \mathbb E[G]  \leq  2 \Re_n(\mathcal L_0).
\end{equation}
Finally, we bound $\Re_n(\mathcal L_0)$ using Dudley's theorem \citep{dudley1987universal}. Conditional on $Z_1, \ldots, Z_n$, Dudley's entropy bound gives
\begin{align}
    \mathbb{E}\left[\sup _{\ell_\theta \in \mathcal L_0} \frac{1}{n} \sum_{i=1}^{n} \mathcal{E}_{i} g \left(Z_{i}\right) \middle| Z_1, \ldots, Z_n\right] &\le \frac{12}{\sqrt{n}}  \int_{0}^{\infty} \sqrt{\log \mathcal N \bigl( \mathcal L_0, \|\cdot\|_{L_2(\hat{\mathbb P}_0)}, \varepsilon \bigr)} \mathrm{d} \varepsilon \notag\\
    &\le \frac{12}{\sqrt{n}} \mathfrak C_{L_2} (\mathcal L_0),\label{eq:dudley}
\end{align}
where $\mathfrak C_{L_2} (\mathcal L_0)$ is the $L^2$ covering entropy integral of the centered loss class
\[ \mathfrak C_{L_2} (\mathcal L_0) = \sup_{Q \in \mathcal P(\mathcal Z)} \int_{0}^{\infty} \sqrt{\log \mathcal N \bigl( \mathcal L_0, \|\cdot\|_{L_2(Q)}, \varepsilon \bigr)} \mathrm{d} \varepsilon.\]
Thus, by taking the expectation of the initial sample over both sides of Equation~\ref{eq:dudley},
\begin{equation}\label{eq:dudley-two}
    \Re_n(\mathcal L_0) \le \frac{12}{\sqrt{n}} \mathfrak C_{L_2} (\mathcal L_0).
\end{equation}

Combining Equations~\ref{eq:excess-risk-emp-process}, \ref{eq:mcdiarmid}, \ref{eq:rademacher} and~\ref{eq:dudley-two},
\begin{equation*}
    \mathscr R(\hat{\mathbb P}_0, \hat{\theta}_0) - \mathscr R_\Theta 
     \leq 
    \frac{24}{\sqrt{n}} \mathfrak{C}_{L_2}(\mathcal L_0)
    + L_\ell \diamZ \sqrt{\frac{\log(1/\delta)}{2n}},
\end{equation*}
with probability over $\hat{\mathbb P}_{0}$ of at least $1 - \delta$.

\end{enumerate}

 \noindent Combining Lemmas~\ref{lemma:excess-risk-classic-1}--\ref{lemma:excess-risk-classic-3} with a union bound yields that $\mathscr R({\mathbb P}, \hat{\theta}_c) - \mathscr R_\Theta$ is upper bounded with probability over $\hat{\mathbb P}_{0}$ of at least $1- 2\delta$ by
\begin{align*}
      \mathscr R({\mathbb P}, \hat{\theta}_c) - \mathscr R_\Theta &= \left( \mathscr R(\mathbb P, \hat{\theta}_c) - \mathscr R(\hat{\mathbb P}_0, \hat{\theta}_c) \right) + \left( \mathscr R(\hat{\mathbb P}_0, \hat{\theta}_c) - \mathscr R(\hat{\mathbb P}_0, \hat{\theta}_0)   \right) + \left( \mathscr R(\hat{\mathbb P}_0, \hat{\theta}_0) - \mathscr R_\Theta \right)
      \\ &\hspace{-2em}\le L_\ell \left(\frac{\log \left(C_a / \delta\right)}{C_{b} n}\right)^{p / d_z} +  \frac{L_\ell (\kappa/\gamma) \left(\frac{m }{n}\right)^{\frac{1}{p}} \diamZ}{1 - L_s} +
      \frac{24}{\sqrt n} \mathfrak C_{L_2}(\mathcal L_0) + L_\ell \diamZ \sqrt{\frac{\log(1/\delta)}{2n}},
\end{align*}
as desired.%
\end{proof}

\subsection{Proof of Theorem~\ref{thm:gen-error-bound-anytime}}

For clarity, we begin by restating Theorem~\ref{thm:gen-error-bound-anytime}:\\

\noindent \textbf{Theorem (Anytime-Valid Generalization Gap)}
{\itshape    Let $\hat \theta_{T}$ be the ERM solution at iteration $T$. If Conditions~\ref{ass-cont-diff-feat-param-Lips-data} and~\ref{cond:lipschitz-sample-adpat} hold, then

    \[    \mathscr R(\mathbb P, \hat{\theta}_T) \leq   \mathscr R(\hat{\mathbb P}_T, \hat{\theta}_T) + L_\ell \left(\frac{\log \left(C_a / \delta\right)}{C_{b} n}\right)^{p / d_z} +    L_\ell S_T(L_s) \left(\frac{m }{n}\right)^{\frac{1}{p}} \diamZ,
\]
        for any $T$, with probability at least $1-\delta$, where $S_T(L_s)$ is the geometric sum defined in Equation~\ref{eq:defn-geo-sum}.%
}\\ 

\begin{proof}
    By the same logic as in the proof of Theorem~\ref{thm:gen-error-bound}, we have 
    \begin{align*}
        | \mathscr R(\mathbb P, \hat{\theta}_T) - \mathscr R(\hat{\mathbb P}_T, \hat{\theta}_T)  | &\le L_\ell W_1(\mathbb P, \hat{\mathbb P}_T) \\
        &\le L_{\ell} \left[ W_1(\mathbb P, \hat{\mathbb P}_0) + W_1 (\hat{\mathbb P_0}, \hat{\mathbb P_T}) \right] \\
        &\le L_{\ell} \left[ W_1(\mathbb P, \hat{\mathbb P}_0) + S_T(L_s)  \left(\frac{m }{n}\right)^{\frac{1}{p}} \diamZ \right] \\
        &\le L_{\ell} \left[ \left(\frac{\log \left(C_a / \delta\right)}{C_{b} n}\right)^{p / d_z} + S_T(L_s)\left(\frac{m }{n}\right)^{\frac{1}{p}} \diamZ \right], 
    \end{align*}
    where the first line follows from Condition~\ref{ass-cont-diff-feat-param-Lips-data} and the Kantorovich--Rubinstein lemma (Lemma~\ref{lemma:kant-rubin}); the third line is an application of Lemma~\ref{lemma:recipr-distortion}; and the fourth line holds with probability over $\hat{\mathbb P}_{0}$ of at least $1 - \delta$ by Lemma~\ref{lemma:wasserstein-conv}. Note that the application of Lemma~\ref{lemma:recipr-distortion} required that the map $\Phi$ (defined in Equation~\ref{eq:defn-closed-loop-sample-adaptation}) is $L_s$-Lipschitz continuous. By the same logic as in previous proofs, this means the above inequalities hold almost surely over reciprocal paths.%
\end{proof}

\subsection{Proof of Corollary~\ref{cor_stop}}

We repeat the corollary for readability:\\

\noindent \textbf{Corollary (Stopping Rule)}
{\itshape    Suppose that Conditions~\ref{ass-cont-diff-feat-param-Lips-data} and \ref{cond:lipschitz-sample-adpat} hold. Let $T$ be a stopping rule (or indeed a possibly data-dependent, random iteration). If we stop reciprocal learning at iteration $T$, the generalization gap bound (Equation~\ref{eq:gen-error-bound-anytime}) holds
    with probability at least $1 - \delta$.
}\\ 

\begin{proof}
    Recall from the proof of Theorem~\ref{thm:gen-error-bound-anytime} that the only step holding with probability at least $1-\delta$ (rather than almost surely) is the bound on $W_p(\mathbb P, \hat{\mathbb P}_0)$ via Lemma~\ref{lemma:wasserstein-conv}. This event concerns the initial sample only. In particular, it does not depend on $T$. The $T$-dependent term in the bound (namely, $L_\ell S_T(L_s) \left(\frac{m }{n}\right)^{\frac{1}{p}} \diamZ$) stems from Lemma~\ref{lemma:recipr-distortion}, which holds pointwise in $\hat{\mathbb P}_0$ and almost surely pathwise. That is, it holds for every $T$ simultaneously along almost every sample path (see the discussion below Lemma~\ref{lemma:recipr-distortion}). Taken together, this means there is an event with probability at least $1-\delta$ on which the generalization gap bound (Equation~\ref{eq:gen-error-bound-anytime}) %
    holds for all $ t \in \mathbb{N}$ simultaneously. That is, the bound is simultaneously valid over all $t \in \mathbb N$, which implies it is also valid for any stopping rule $T$ (or, in fact, any random $T$).
\end{proof}

\subsection{Proof of Theorem~\ref{thm:excess-risk-classic-anytime}}

We begin by restating Theorem~\ref{thm:excess-risk-classic-anytime} for the reader's convenience:\\

\noindent \textbf{Theorem (Anytime-Valid Excess Risk Bound)}
{\itshape    Assume that $\Theta$ is closed and convex. %
    Given Conditions~\ref{ass-cont-diff-feat-param},~\ref{ass-str-convex} and~\ref{cond:lipschitz-sample-adpat}, the excess risk
    $ \mathscr R({\mathbb P}, \hat{\theta}_T) - \mathscr R_\Theta $ is upper bounded by
    \[
         L_\ell \left(\frac{\log \left(C_a / \delta\right)}{C_{b} n}\right)^{p / d_z}
        + L_\ell  S_T(L_s) \frac{\kappa}{\gamma} \left(\frac{m }{n}\right)^{\frac{1}{p}} \diamZ
        + \frac{24}{\sqrt n} \mathfrak C_{L_2}(\mathcal L_0) + L_\ell \diamZ \sqrt{\frac{\log(1/\delta)}{2n}},
    \]
    for any stopping rule $T$, with probability at least $1-2\delta$.%
}\\

\begin{proof}
As with the proof of Theorem~\ref{thm:excess-risk-classic}, there are three lemmas which are combined using a union bound argument. We begin by proving each lemma in turn: 
\begin{enumerate}[label=(\Alph*)]
    \item\label{proof:excess-risk-classic-anytime-1} By replacing $\hat{\theta}_c$ by $\hat{\theta}_T$ throughout Part~\ref{proof:excess-risk-classic-1} of Theorem~\ref{thm:excess-risk-classic}'s proof, we have that 
    \begin{equation*}
        \mathscr R({\mathbb P}, \hat{\theta}_T) - \mathscr R(\hat{\mathbb P}_0, \hat{\theta}_T) \leq   L_\ell \left(\frac{\log \left(C_a / \delta\right)}{C_{b} n}\right)^{p / d_z},
    \end{equation*}
    with probability over $\hat{\mathbb P}_{0}$ of at least $1 - \delta$ (and surely pathwise). 
    \item\label{proof:excess-risk-classic-anytime-2} It is easy to see that Equation~\ref{eq:theta-distortion} in Part~\ref{proof:excess-risk-classic-2} of the proof of Theorem~\ref{thm:excess-risk-classic} holds for every $T$. Its derivation uses neither Condition~\ref{cond:conv-reciprocal-learning} nor the assumption $L_s < 1$ from Theorem~\ref{thm:excess-risk-classic}, so we can apply it here. Since $\mathscr R(\hat{\mathbb P}_0,\, \cdot\,)$ is $L_\ell$-Lipschitz continuous by Condition~\ref{ass-cont-diff-feat-param-cts-diff-param}, %
    it follows that
    \begin{equation}\label{eq:proof:excess-risk-classic-anytime-2-bound}
        \mathscr R(\hat{\mathbb P}_0, \hat{\theta}_T) - \mathscr R(\hat{\mathbb P}_0, \hat{\theta}_0)  \leq L_\ell \|\hat{\theta}_0 - \hat{\theta}_{T}\|_2 \leq L_\ell L_a S_T(L_s) \left(\frac{m }{n}\right)^{\frac{1}{p}} \diamZ,
    \end{equation}
    where $L_a = \kappa/\gamma$ is a Lipschitz constant for the ERM map $\hat \theta (P) := \argmin_{\theta\in\Theta}\mathscr R(P,\theta)$. (That this map is well defined is proven in Part~\ref{proof:excess-risk-classic-2} of Theorem~\ref{thm:excess-risk-classic}'s proof.) Equation~\ref{eq:proof:excess-risk-classic-anytime-2-bound} holds whenever the map $\Phi$ (defined in Equation~\ref{eq:defn-closed-loop-sample-adaptation}) is $L_s$-Lipschitz continuous. By the same reasoning as in previous proofs, this means Equation~\ref{eq:proof:excess-risk-classic-anytime-2-bound} holds almost surely pathwise.

    \item\label{proof:excess-risk-classic-anytime-3} Lemma~\ref{proof:excess-risk-classic-3} in proof of Theorem~\ref{thm:excess-risk-classic} shows that 
    \begin{equation*}
        \mathscr R(\hat{\mathbb P}_0, \hat{\theta}_0) - \mathscr R_\Theta 
        \leq 
        \frac{24}{\sqrt{n}} \mathfrak{C}_{L_2}(\mathcal L_0)
        + L_\ell \diamZ \sqrt{\frac{\log(1/\delta)}{2n}} ,
    \end{equation*}
    with probability over $\hat{\mathbb P}_{0}$ of at least $1 - \delta$ (and surely pathwise).

\end{enumerate}

\noindent The claim then follows by combining the above three lemmas using a union bound argument, as we did in the proof of Theorem~\ref{thm:excess-risk-classic}. %
\end{proof}

\subsection{Proof of Theorem~\ref{thm:data-dependent-anytime-excess}}

To facilitate exposition, we restate Theorem~\ref{thm:data-dependent-anytime-excess}:\\

\noindent \textbf{Theorem (Data-Dependent Anytime-Valid Excess Risk Bound)}
{\itshape    Given Condition~\ref{ass-cont-diff-feat-param-Lips-data}, the excess risk
    $ \mathscr R({\mathbb P}, \hat{\theta}_T) - \mathscr R_\Theta $ is upper bounded by
    \[
        \mathscr R(\hat{\mathbb P}_0, \hat{\theta}_T) - \mathscr R(\hat{\mathbb P}_0, \hat{\theta}_0) + L_\ell \left(\frac{\log \left(C_a / \delta\right)}{C_{b} n}\right)^{p / d_z}
        +
        \frac{24}{\sqrt n} \mathfrak C_{L_2}(\mathcal L_0) + L_\ell \diamZ \sqrt{\frac{\log(1/\delta)}{2n}},
    \]
    for any stopping rule $T$, with probability at least $1-2\delta$.%
}\\ 

\begin{proof}
    This proof %
    is the same as that of Theorem~\ref{thm:excess-risk-classic-anytime}, except the bound given by Equation~\ref{eq:proof:excess-risk-classic-anytime-2-bound} is not used. In its place, we keep the term $\mathscr R(\hat{\mathbb P}_0, \hat{\theta}_T) - \mathscr R(\hat{\mathbb P}_0, \hat{\theta}_0)$. Hence, Theorem~\ref{thm:data-dependent-anytime-excess} holds surely over reciprocal paths, since only Lemma~\ref{proof:excess-risk-classic-anytime-2} in Theorem \ref{thm:excess-risk-classic-anytime}'s proof required the map $\Phi$ (defined in Equation~\ref{eq:defn-closed-loop-sample-adaptation}) to be Lipschitz continuous.
\end{proof}

\subsection{Proof of Proposition~\ref{prop:boosting}}\label{sec:proof-boosting}

We restate the proposition before proving it for clarity:\\

\noindent \textbf{Proposition (Improved Bounds for Boosting)}
{\itshape    Assume that reciprocal learning is non-greedy and modifies $m_{t+1}$ of the $n_t$ %
    sample points when transitioning from $\hat{\mathbb P}_t$ to $\hat{\mathbb P}_{t+1}$.  %
    Further, suppose
    that every modified point keeps its covariates fixed and only changes its labels (as in boosting, for example).
    Then Lemma~\ref{lemma:recipr-distortion} holds with $\diamZ$ replaced by
    $
    \diamY:=\sup_{y,y'\in\mathcal Y} \| y - y'\|_2 <\infty.
    $
    Hence, the generalization gap bound (Theorem~\ref{thm:gen-error-bound}),
    the excess risk bound (Theorem~\ref{thm:excess-risk-classic}), the anytime-valid generalization gap bound (Theorem~\ref{thm:gen-error-bound-anytime}), and the anytime-valid excess risk bound (Theorem~\ref{thm:excess-risk-classic-anytime}) can all be tightened
    by replacing %
    the term $\left(\frac{m}{n}\right)^{\frac{1}{p}} \diamZ$ with $\left(\frac{m}{n}\right)^{\frac{1}{p}}\diamY$. %
}\\ 

\begin{proof}
The proof works by modifying the only step at which $\mathcal D_{\mathcal Z}$ enters the
proof of the results we are tightening---namely, Equation~\ref{eq:diameter} in the proof of
Lemma~\ref{lemma:recipr-distortion}. %
Fix an iteration $t$ and consider
$
\hat{\mathbb P}_t=\frac1n\sum_{i=1}^n \delta_{Z_i^{(t)}}$ %
and $
\hat{\mathbb P}_{t+1}=\frac1n\sum_{i=1}^n \delta_{Z_i^{(t+1)}}
$. (In this setting, reciprocal learning is non-greedy because no samples are added, only existing ones are modified; hence $n_t = n_{t+1} = n$.)
By assumption, there exists an index set $I_t\subseteq\{1,\dots,n\}$ with $|I_t|= m_{t+1}$ %
such that
$
Z_i^{(t+1)}=Z_i^{(t)} \text{ for all } i\notin I_t,
$
and
$
Z_i^{(t+1)}=(X_i^{(t)},\tilde Y_i^{(t+1)}) \text{ for all } i\in I_t,
$
where $\tilde Y_i^{(t+1)}$ denotes a modified label. 

Now let $c_t$ denote the coupling of $\hat{\mathbb P}_t$ and
$\hat{\mathbb P}_{t+1}$ given by
\[
c_t:=\frac1n\sum_{i=1}^n \delta_{(Z_i^{(t)},Z_i^{(t+1)})}.
\]
By the definition of the $p$-Wasserstein distance (Definition~\ref{def:wasserstein}) as an infimum over all couplings,
\[
W_p^p(\hat{\mathbb P}_t,\hat{\mathbb P}_{t+1})
\le \int_{\mathcal Z\times\mathcal Z} \| z - z' \|_2^p  \mathrm{d} c_{t}(z,z')
= \frac{1}{n}\sum_{i=1}^{n} \bigl\| Z_i^{(t)} - Z_i^{(t+1)} \bigr\|_2^p .
\]
For $i\notin I_t$ the summands vanish. For $i\in I_t$, the covariate $X_i^{(t)}$ is unchanged and hence,
\[
\|Z_i^{(t)} - Z_i^{(t+1)}\|_2
=
\|(X_i^{(t)},Y_i^{(t)}) - (X_i^{(t)},\tilde Y_i^{(t+1)})\|_2
= \| Y_i^{(t)} - \tilde Y_i^{(t+1)}\|_2
\le \diamY.
\]
Therefore,
\[
W_p^p(\hat{\mathbb P}_t,\hat{\mathbb P}_{t+1})
\le \frac{|I_t|}{n}\diamY^p
= \frac{m_{t+1}}{n}\diamY^p.
\]
That is,
\[
W_p(\hat{\mathbb P}_t,\hat{\mathbb P}_{t+1})
\le \left(\frac{m_{t+1}}{n}\right)^{1/p}\diamY.
\]
In particular, for $t=0$, this replaces Equation~\ref{eq:diameter} in the proof of Lemma~\ref{lemma:recipr-distortion} by
\[
W_p\bigl(f_s(\hat\theta_0,\hat{\mathbb P}_0),\hat{\mathbb P}_0\bigr)
\le \left(\frac{m}{n}\right)^{1/p}\diamY.
\]
Now observe that the remainder of the proof of Lemma~\ref{lemma:recipr-distortion} is unchanged. Hence,
\begin{equation}\label{eq:sharper-lemma-recipr-distortion}
    W_p(\hat{\mathbb P}_0,\hat{\mathbb P}_T) \le S_T(L_s)\left(\frac{m}{n}\right)^{1/p}\diamY.
\end{equation}
Equation~\ref{eq:sharper-lemma-recipr-distortion} is a sharper version of Lemma~\ref{lemma:recipr-distortion}. We can directly substitute it into the places Lemma~\ref{lemma:recipr-distortion} is used in the proofs of Theorems~\ref{thm:gen-error-bound}, \ref{thm:excess-risk-classic}, \ref{thm:gen-error-bound-anytime} and~\ref{thm:excess-risk-classic-anytime}.
This immediately gives tighter bounds in all four theorems,
with $\left(\frac{m}{n}\right)^{1/p}\diamZ$ replaced by $\left(\frac{m}{n}\right)^{1/p}\diamY$.
\end{proof}

\section{Historical Background on Feedback-Driven Learning}\label{app-history}
\begin{footnotesize}
    \begin{quote}
        \say{Artificial intelligence was born at a conference at Dartmouth in 1956 [...], three years after the Macy conferences on cybernetics had ended. The two movements coexisted for roughly a decade, but by the mid-1960s, the proponents of symbolic AI gained control of national funding conduits and ruthlessly defunded cybernetics research. This effectively liquidated the subfields of \emph{self-organizing} systems, neural networks and \emph{adaptive} machines [...]}      \flushright -- \citet{cariani2010importance}, page 89 [emphasis added].
    \end{quote}
\end{footnotesize}
Eight years before the famous Dartmouth conference on AI, \citet{wiener1948cybernetics} established cybernetics as the interdisciplinary study of \say{control and communication in the animal and the machine.} In the Macy conferences from 1941 to 1960, cybernetics was defined 
as the study of ``circular causal and feedback mechanisms'' \citep{von1952cybernetics}.
While AI focused on \textit{static} data-driven pattern recognition at that time, cybernetics research addressed the \textit{dynamic} principles of system stability, adaptability, and (feedback-driven) control theory, encompassing both artificial and biological systems. 
In the years that followed, AI research by and large flourished, while cybernetics remained a niche discipline. 

Six decades later, it is fair to say that ideas from the realm of applied cybernetics---most notably, artificial neural networks---have made it back to the AI mainstream. (For a summary of early deliberations on neural networks by the cybernetics community, see \citet{von2003self}.) Surprisingly, the bulk of modern (machine) learning theory, however, still takes a mostly static point of view, relying on single-step ERM \citep{vapnik1968uniform,vapnik1991principles,vapnik1998statistical,cherkassky2015measures}. This is all the more surprising as their founding fathers, Vladimir Vapnik and Alexej Chervonenkis, were both students of Alexander Lerner, a leading researcher in cybernetics. In fact, their seminal paper \say{The uniform convergence of frequencies of the appearance of events to their probabilities} \citep{vapnik1968uniform} was published in the \textit{Control Sciences} division of the Proceedings of the Academy of Science, after being declined for publication in the \textit{Statistics} division \citep[see][for the full story]{Obituary}. 

Excitingly, theoretical machine learning has seen some revival of ideas around system stability, adaptability, and feedback-driven control in recent years. The resulting developments have, so far, much less significance than those by \citet{vapnik1968uniform}, but the resurgence of cybernetics and control theory is captivating on its own terms, as detailed in the introduction.

\end{document}